\documentclass[10pt,twocolumn,letterpaper]{article}

\usepackage[pagenumbers]{cvpr} % To force page numbers, e.g. for an arXiv version

\usepackage[dvipsnames]{xcolor}

\definecolor{cvprblue}{rgb}{0.21,0.49,0.74}

\usepackage{subcaption}
\usepackage{graphicx}
\usepackage{amsmath}
\usepackage{amssymb}
\usepackage{booktabs}
\usepackage{adjustbox}
\usepackage{makecell}
\usepackage{multirow}
\usepackage{bbding}
\usepackage{xcolor}
\usepackage{color, colortbl}
\definecolor{citecolor}{HTML}{2980b9}
\definecolor{linkcolor}{HTML}{c0392b}

\usepackage{enumerate}
\usepackage{bbding}
\usepackage{pifont}
\usepackage{wasysym}
\usepackage{float}

\usepackage{algorithm}
\usepackage{algorithmicx}
\usepackage{algpseudocode}
\newcommand{\ms}[1]{\boldsymbol{#1}}

\usepackage[hidelinks,pagebackref=true,breaklinks=true,colorlinks,bookmarks=false,citecolor=citecolor,letterpaper=true,linkcolor=linkcolor]{hyperref}

\makeatletter
\newcommand\figcaption{\def\@captype{figure}\caption}
\newcommand\tabcaption{\def\@captype{table}\caption}
\makeatother

\usepackage{soul}

\newcommand{\lin}[1]{\textcolor{black}{#1}}

\newcommand{\rred}[1]{\textcolor[RGB]{197,74,61}{#1}}

\def \W{{\bf W}}

\definecolor{Redref}{RGB}{224,156,151}

\definecolor{ForestGreen}{RGB}{34,139,34}
\definecolor{brickred}{rgb}{0.8, 0.25, 0.33}

\usepackage{wasysym}

\usepackage{cellspace}
\usepackage{mathtools}
\usepackage[thinc]{esdiff}
\makeatletter
\def\adl@drawiv#1#2#3{%
        \hskip.5\tabcolsep
        \xleaders#3{#2.5\@tempdimb #1{1}#2.5\@tempdimb}%
                #2\z@ plus1fil minus1fil\relax
        \hskip.5\tabcolsep}
\newcommand{\cdashlinelr}[1]{%
  \noalign{\vskip\aboverulesep
          \global\let\@dashdrawstore\adl@draw
          \global\let\adl@draw\adl@drawiv}
  \cdashline{#1}
  \noalign{\global\let\adl@draw\@dashdrawstore
          \vskip\belowrulesep}}
\makeatother

\def\paperID{15459} % *** Enter the Paper ID here
\def\confName{CVPR}
\def\confYear{2024}

\title{MoPE-CLIP: Structured Pruning for Efficient Vision-Language Models with Module-wise Pruning Error Metric}

\author{
\textbf{Haokun Lin$^{1,2,4}$,  ~ Haoli Bai$^3$,  ~ Zhili Liu$^{3,5}$,  ~ Lu Hou$^3$,  ~ Muyi Sun$^2$,} \\
\textbf{Linqi Song$^4$, ~    
Ying Wei$^{6}$\footnotemark[2], ~
and Zhenan Sun$^2$}\footnotemark[2] \\
$^1$ School of Artificial Intelligence, University of Chinese Academy of Sciences\\
$^2$ CRIPAC \& MAIS, Institute of Automation, Chinese Academy of Sciences\\
$^3$Huawei Noah's Ark Lab \quad$^4$City University of Hong Kong \\
$^5$ The Hong Kong University of Science and Technology \quad$^6$ Nanyang Technological University\\
\texttt{haokun.lin@cripac.ia.ac.cn,ying.wei@ntu.edu.sg,znsun@nlpr.ia.ac.cn}}

\begin{document}
\maketitle
\renewcommand{\thefootnote}{\fnsymbol{footnote}}
\footnotetext[2]{Corresponding authors.}
\begin{abstract}
% The ABSTRACT is to be in fully justified italicized text, at the top of the left-hand column, below the author and affiliation information.
% Use the word ``Abstract'' as the title, in 12-point Times, boldface type, centered relative to the column, initially capitalized.
% The abstract is to be in 10-point, single-spaced type.
% Leave two blank lines after the Abstract, then begin the main text.
% Look at previous \confName abstracts to get a feel for style and length.
% % 

Vision-language pre-trained models have achieved impressive performance on various downstream tasks.
However, their large model sizes hinder their utilization on platforms with limited computational resources.
We find that directly using smaller pre-trained models and applying magnitude-based pruning on CLIP models leads to inflexibility and inferior performance.
Recent efforts for VLP compression either adopt uni-modal compression metrics resulting in limited performance or involve costly mask-search processes with learnable masks.
In this paper, we first propose the Module-wise Pruning Error (MoPE) metric, accurately assessing CLIP module importance by performance decline on cross-modal tasks.
Using the MoPE metric, we introduce a unified pruning framework applicable to both pre-training and task-specific fine-tuning compression stages. 
For pre-training, MoPE-CLIP effectively leverages knowledge from the teacher model, significantly reducing pre-training costs while maintaining strong zero-shot capabilities.
For fine-tuning, consecutive pruning from width to depth yields highly competitive task-specific models.
Extensive experiments in two stages demonstrate the effectiveness of the MoPE metric, and MoPE-CLIP outperforms previous state-of-the-art VLP compression methods.

\end{abstract}    
\section{Introduction}
\label{sec:intro}

\begin{figure*}[!ht]
    \centering
    \includegraphics[width=1.\linewidth]{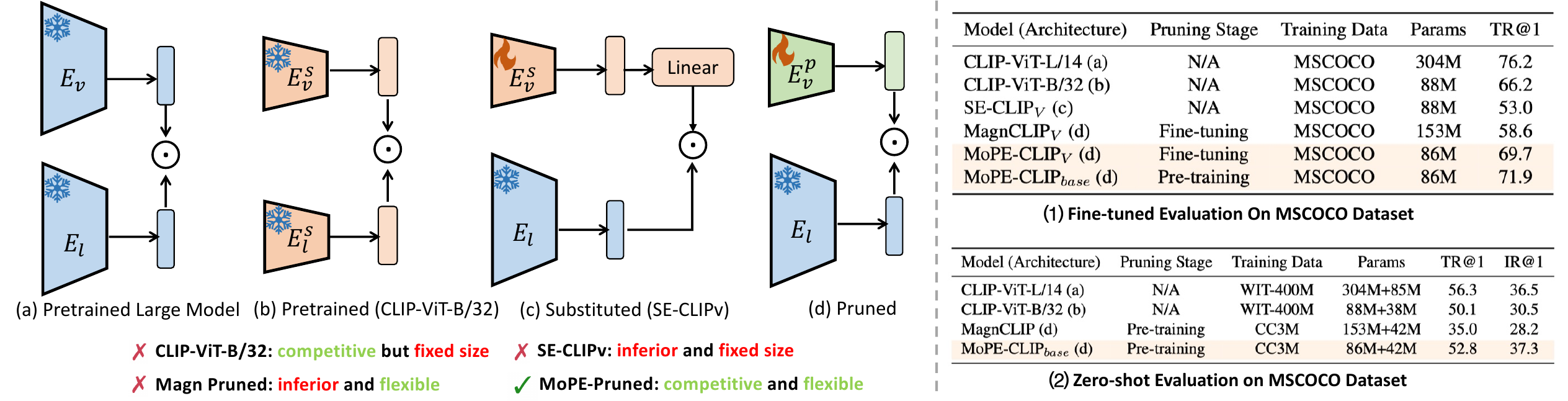}    
    \caption{
    Empirical comparison between (a) the original large CLIP model and three smaller models with compressed vision encoders, including 
    (b) a \textbf{pre-trained} small CLIP Model;
    (c) a small model obtained by \textbf{substituting} the original vision encoder in (a) with the small vision encoder of (b); and
    (d) a small model with the vision encoder  \textbf{pruned} from  (a).
    We perform pruning during pre-training or fine-tuning, evaluated (1) after fine-tuning or (2) with zero shot.
    Note that we train the substituted encoder $E^s_v$ in (c) and $E_v^p$ in (d) with image-text contrastive loss $\mathcal{L}_{itc}$.
    TR and IR stand for image-to-text and text-to-image retrieval, respectively.
    }
    \vspace{-1.em}
    \label{fig:small-models}    
\end{figure*}

Vision-Language Pre-training (VLP) models have demonstrated strong multi-modal representation learning abilities \citep{kim2021vilt,jia2021ALIGN,li2023filp_km,li2023blip_v2}. 
However, their impressive performance comes at the cost of a large number of parameters, limiting their 
% \yingtao{deployment} 
use on resource-constrained devices.
Therefore, it is essential to explore compact VLP models for real-world applications \citep{wang2022efficientvlm, shi2023upop}.
We identify two compression settings for different platforms. 
First, many edge servers lack the computational power to handle the entire pre-trained model. 
% To address this limitation while maintaining versatility across tasks, 
We define ``\textbf{pre-training stage compression}" to address this,
which involves compressing zero-shot VLP models and pre-training them on millions of image-text pairs to create compact, general-purpose models.
Second, clients, like mobile phones, often require multiple task-specific models for various scenarios. 
To meet this demand, we introduce ``\textbf{fine-tuning stage compression}". 
For example, CLIP \citep{radford2021clip} excels in cross-modal retrieval tasks, comprising image-to-text retrieval (TR) and text-to-image retrieval (IR). 
% Leveraging pre-computable and offline-storable visual or textual representations \citep{yao2021filip, dai2022enabling}, we aim to compress the vision encoder for the TR task and the text encoder for the IR task.
Given the pre-computable and offline storable nature of visual or textual representations \citep{yao2021filip, dai2022enabling}, our objective is to compress vision encoder for TR task and text encoder for IR task.

To reduce inference costs, smaller pre-trained models, like the various-sized ViT-based CLIP models in \citep{radford2021clip}, are considered.
However, individually pre-training each model is computationally expensive \citep{cherti2023openclip}, and the limited architectural diversity may not meet various deployment needs. 
Consequently, we delve into more flexible solutions that leverage pruning techniques to compress VLP models. Nonetheless, the suboptimal performance of magnitude pruning on CLIP as shown in Figure \ref{fig:small-models} raises the challenge of identifying a more competitive pruning strategy.

Recent VLP pruning methods \citep{wang2022efficientvlm, shi2023upop, wu2023tinyclip} can be broadly categorized into two categories. 
The simplest way involves applying uni-modal Transformer pruning methods. 
However, despite the effectiveness of metrics such as magnitude and loss-awareness on single-modality transformers \citep{han2015learning,he2018soft,molchanov2019importance_estimation,michel2019sixteenheads}, our experiments have revealed unsatisfactory performance when directly applying them to the multimodal CLIP models.
EfficientVLM \citep{wang2022efficientvlm} uses the ``every other" pruning strategy during the pre-training stage, but this approach, commonly used in BERT models, does not deliver optimal outcomes in our experiments.
These findings underscore the inadequacy of existing metrics in assessing module impact on multi-modal tasks.
On the other hand,  mask-based pruning is employed to identify crucial modules. 
UPop \citep{shi2023upop} introduces a progressive searching process, which is unsuitable for the pre-training stage.
TinyCLIP \citep{wu2023tinyclip} suggests distillation with weight inheritance for small models, involving a time-consuming multi-stage distillation process. 
Additionally, TinyCLIP is pre-trained on the LAION400M dataset \citep{schuhmann2021laion}, leaving uncertainty about its effectiveness for fine-tuning stage compression with limited data.
In summary, traditional pruning metrics need improvement for VLP pruning, and mask-based pruning is not efficient enough during pre-training. 
Thus, a unified solution for our identified two-stage compression is unexplored.

To tackle these challenges, we introduce \textbf{MoPE-CLIP}, an effective mask-free structured pruning solution for both pre-training and fine-tuning stage compression.
We first propose the Module-wise Pruning Error (MoPE) metric, which quantifies a module's importance by measuring the performance drop in multi-modal tasks if that module is pruned. 
MoPE precisely evaluates the pruning sensitivity of heads, FFN neurons in the \textit{width direction}, and Transformer layers in the \textit{depth direction}.
Based on the MoPE metric, we propose a unified mask-free pruning framework.
In the pre-training stage, we calculate MoPE using zero-shot retrieval on the MSCOCO validation set and simultaneously prune both width and depth components.
In the fine-tuning stage, 
MoPE is calculated by the performance decline on downstream tasks.
To achieve higher pruning ratios,
we prioritize pruning in the width direction before pruning in the depth direction.
Moreover, 
we distill both \textit{cross-modal} and \textit{uni-modal} knowledge from the original model's aligned feature space and text/vision encoder to enhance the pruned model's capacity.
Extensive evaluations demonstrate our MoPE-CLIP largely outperforms the same amount of parameters TinyCLIP
\citep{wu2023tinyclip} by 5.3\% TR@1 and 4.0\% IR@1 on MSCOCO retrieval tasks, while surpassing MCD \citep{kim2023mcd} and ALIP \citep{yang2023alip} on 11 zero-shot classification tasks by 18.6\% and 17.0\%.
The contributions of our work are:

\begin{itemize}
    \item We introduce MoPE metric for precisely assessing the importance of CLIP modules in cross-modal tasks. Utilizing MoPE, we present a structured pruning framework combined with advanced distillation loss, offering a unified solution for pre-training and fine-tuning compression.
    \item MoPE-CLIP model exhibits SOTA performance in both training speed and accuracy across extensive experiments, surpassing existing benchmarks in various domains.
    \vspace{-0.1cm}
\end{itemize}

\section{Preliminary Study of Downsizing~CLIP}

% TODO: Revise the figures 
\begin{figure*}[!ht]
    \centering
    \includegraphics[width=0.95\linewidth]{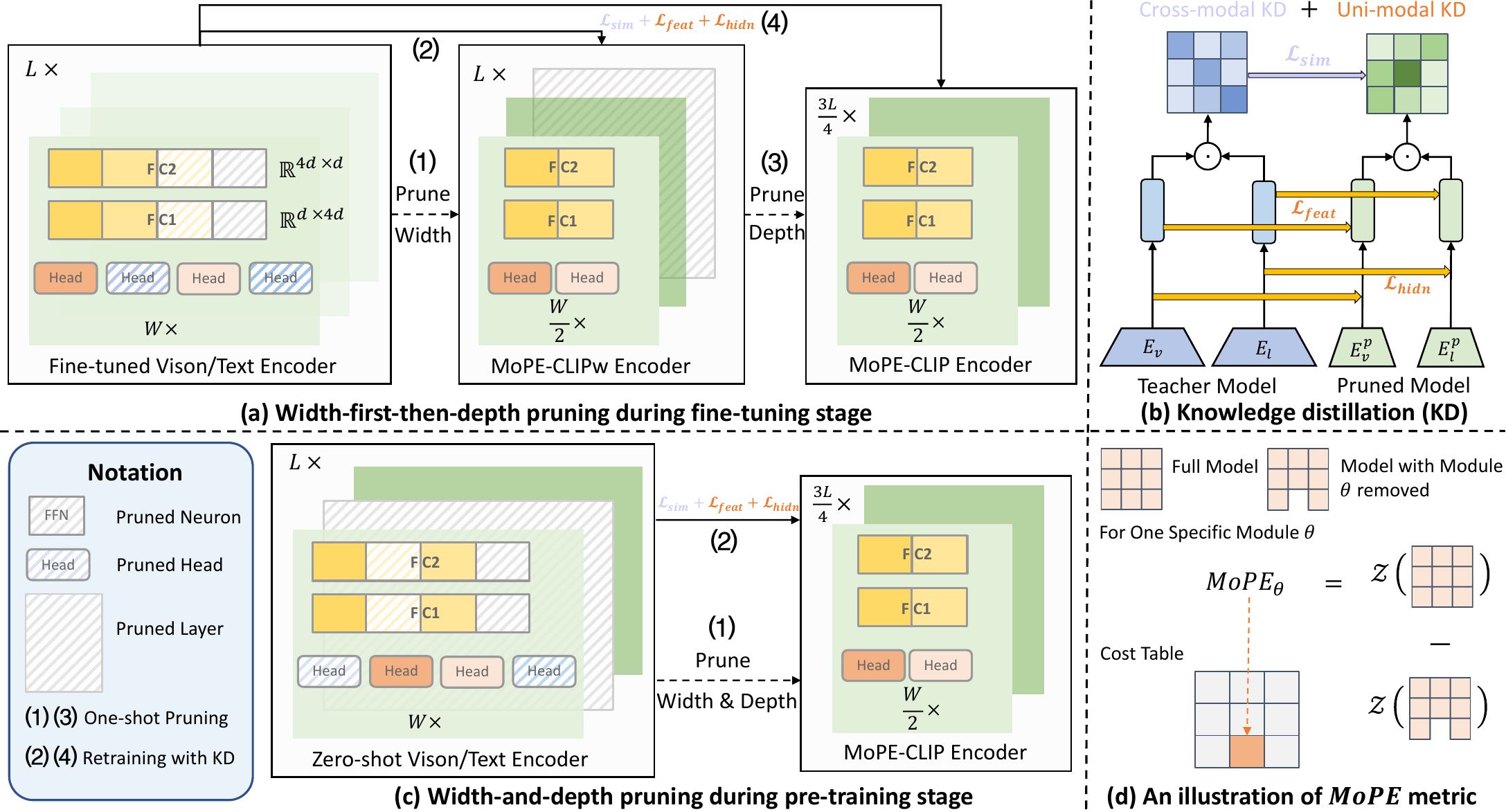}
    \caption{
    The overall workflow of training MoPE-CLIP. 
    (a). During the fine-tuning stage, we apply width-first-then-depth pruning on fine-tuned CLIP vision or text encoder to obtain powerful task-specific models. 
    (b). An illustration of our distillation process, transferring cross-modal and uni-modal knowledge.
    (c). During the pre-training stage,  we apply consecutive pruning in the width and depth directions on zero-shot CLIP encoders.
    (d). An illustration of MoPE metric, measuring the performance drop of CLIP after removing the module $\theta$.
    }
    \vspace{-0.7em}
    \label{fig:pruning-process}    
\end{figure*}

\label{sec:preliminary}

In pursuit of the objective to scale down a vision-language model such as CLIP, various alternatives come into consideration. Architecturally, one may opt to substitute an encoder with its counterpart in a smaller model, as exemplified by CLIP-ViT-B/32 in Figure~\ref{fig:small-models}(b), or alternatively, directly prune the encoder to any desired size. From a practical standpoint, downsizing can be executed either during pre-training prior to deployment for downstream tasks or during fine-tuning in clients. 
% Both find relevance in diverse contexts, as discussed in the Introduction. 
This section embarks on a preliminary examination of these alternatives, laying the groundwork for our proposed pruning strategy.

\vspace{-0.3cm}

\paragraph{Substituting with smaller models proves unsatisfactory.}
We substitute the original vision encoder $E_v$ of CLIP with a smaller one from CLIP-ViT-B/32, resulting in the downsized model SE-CLIP$_V$. We freeze the language encoder to facilitate applications like image-to-text retrieval (TR), where text features by the language encoder are oftentimes stored without modification.
The modified vision encoder and the frozen language encoder are misaligned, necessitating further training. Concretely, we conduct fine-tuning of a linear layer and the vision encoder on the downstream dataset \lin{MSCOCO \citep{lin2014mscoco}}, and resort to the cross-modal contrastive loss function  \lin{$\mathcal{L}_{itc}$, which is the InfoNCE loss computed between image and text features.}
Unfortunately, SE-CLIP$_V$ experiences a substantial performance decline compared with the original CLIP, as illustrated in Figure~\ref{fig:small-models}(1). This decline may be attributed to the formidable challenge of aligning two disassociated encoders originating from distinct vision-language models. This observation, coupled with the lack of flexibility in selecting a target size, dissuades us from further investigating this downsizing strategy during pre-training. Thus, we redirect our focus to the alternative choice of pruning.

\vspace{-0.5cm}

\paragraph{Further investigation is required for successful pruning.} 
Specifically, we implement MagnCLIP$_V$~\citep{han2015learning}, a widely adopted yet straightforward pruning strategy, which selectively prunes 
% XXX [layers? heads? or specific neurons?] 
attention heads and FFN neurons
below a specified magnitude threshold. 
Adjusting this threshold results in varying sizes of pruned models. 
We utilize the same objective function as $\mathcal{L}_{itc}$
% XXX 
in substitution to train the pruned model $E_v^p$. Despite the expected flexibility in target size, MagnCLIP$_V$ ensures only a relatively satisfactory performance, provided that at least $50\%$ of parameters are retained, as depicted in Figure~\ref{fig:small-models}(1)(2). An imperative need exists for an effective pruning strategy that simultaneously meets flexibility and generalization capacity, thus forming the basis for our proposed approach.
\vspace{-1em}

\paragraph{Both pre-training and fine-tuning pruning merit consideration.}
An intriguing question to explore is whether a vision-language model, pre-trained before deployment, outperforms one pruned to the same size during fine-tuning.
MoPE-CLIP$_{base}$ and MoPE-CLIP$_{V}$ represent the two versions using our proposed
% unified
pruning framework
which we will detail in the next. 
From Figure~\ref{fig:small-models}(1), we conclude that given the same target size, exploring both pre-training pruning and fine-tuning pruning is worthwhile. First, their application scenarios differ as discussed. Second, pruning during pre-training, when more parallel data is accessible for preserving cross-modal correlations, proves more effective, while pruning during fine-tuning which does not underperform significantly enjoys the advantage of high efficiency.

\section{Method}
We introduce the MoPE metric in Section \ref{subsec:method_MoPE} to measure module influence accurately in cross-modal tasks. 
In Sections \ref{subsec:method_MoPEclip} and \ref{subsec:method_kd}, we present our pruning framework and knowledge distillation loss, jointly improving the two-stage compression performance.

\subsection{Module-wise Pruning Error} \label{subsec:method_MoPE}

\lin{We propose a new metric called \textit{module-wise pruning error} (MoPE) to evaluate the importance of different modules in the CLIP model, such as Multi-Head Attention (MHA) heads, Feedforward (FFN) neurons, and entire Transformer layers.}
For heads and neurons, some commonly used metrics in width pruning, like magnitude \citep{han2015learning}, 
fail to accurately capture their impacts on multi-modal tasks, leading to suboptimal results.
% cannot correctly reflect the impact of these modules on multi-modal tasks.
For Transformer layers, 
existing works \citep{sajjad2020poormanbert, fan2019reducingbertdepth} mainly adopt every other strategy on depth pruning for BERT. 
% without a comprehensive evaluation of the importance between layers. 
\lin{Our experiments in Section \ref{subsec:ablation} demonstrate that this simplistic strategy falls short when applied to CLIP models.
We suppose that every other strategy cannot measure the pruning influence on another encoder and thus leads to inferior performance.
}
These results present a new challenge in selecting a suitable metric to prune VLP models.

%TODO: Use new notation and equation to illustrate this metric

To overcome these issues, our proposed MoPE metric effectively assesses the module's importance with respect to multi-modal downstream tasks, offering a consistent and more accurate measure for both width and depth pruning. 
% The MoPE metric proves to be more accurate and efficient when applied to VLP models.
In particular, we regard different heads, groups of neurons, and layers as different modules.
From Figure~\ref{fig:pruning-process} (d), the importance of module $\theta$ is empirically measured by the performance decline between module $\theta$ removed CLIP model $f_{\varphi-\theta}$  and to the full CLIP $f_{\varphi }$  counterpart as follows:
\begin{equation} \label{eq:MoPE}
    {\rm MoPE}_{\theta} = \mathcal{Z}\left [ f_{\varphi}\right ] - \mathcal{Z}\left [f_{\varphi-\theta} \right ],
% \vspace{-0.5em}
    \vspace{-5pt}
\end{equation}
where $\mathcal{Z}$
is the zero-shot evaluation function, i.e., Recall Mean for retrieval tasks.
% One module with 
One module $\theta$ with a higher MoPE$_{\theta}$ value indicates that this module is more sensitive to pruning and plays a more crucial role in cross-modal tasks.
% By utilizing the MoPE$_{\theta}$, we can easily create cost tables 
% $\mathcal{C}_{\theta} = \sum_{i=1}^{n}\left\{ \text{MoPE}_{\theta_{i}} \right\}$, and preserve these modules with higher MoPE$_{\theta}$ value.
Thus, preserving such modules becomes a priority during pruning. 
By utilizing the MoPE$_{\theta}$, we can easily create cost tables 
$\mathcal{C}_{\theta} = \sum_{i=1}^{n}\left\{ \text{MoPE}_{\theta_{i}} \right\}$.
These cost tables are generated for different heads ($\mathcal{C}_{head}$), groups of neurons ($\mathcal{C}_{neuron}$), and layers ($\mathcal{C}_{layer}$).
They serve as references for selecting optimal architectures, allowing us to retain critical modules while reducing the overall model size.
% The cost tables serve as a reference point for selecting optimal architectures, allowing us to retain critical modules while reducing the overall model size.

\subsection{Unified Pruning Framework Based on MoPE} 
\label{subsec:method_MoPEclip}
\lin{Recent VLM compression works focus either during the pre-training stage \citep{wu2023tinyclip} or fine-tuning stage \citep{shi2023upop}.
However, the comprehensive solution for these two stages is under-explored.} 
% Based on our MoPE metric, we propose a unified pruning framework termed this challenge.
Leveraging our MoPE metric, we introduce a unified pruning framework aimed at solving this challenge.

\paragraph{Fine-tuning Stage.}

\vspace{-0.3cm}
The primary challenge lies in enhancing the performance of task-specific pruned models. 
To achieve high compression ratios, we explore three distinct pruning strategies in both width and depth directions. Empirical analysis reveals that the \textbf{width-first-then-depth} pruning paradigm is the most effective, as discussed in Section \ref{subsec:ablation}.
% Specifically, one encoder of CLIP has $L$ layers, and we first compress all layers of fine-tuned CLIP in the width direction, as shown in Figure \ref{fig:pruning-process} (b).
Specifically, one encoder of CLIP has $L$ layers, and each layer consists of a MHA block and a FFN block.
% Specifically, one encoder of CLIP has $L$ layers.
First, we compress the fine-tuned CLIP model in the width direction, as shown in Figure \ref{fig:pruning-process} (a).
For the MHA block with $N_{h}$ heads, we independently prune $L \times N_{h}$ heads with their query, key, value, and output matrices. This process calculates the MoPE metric, subsequently establishing the cost table of heads $\mathcal{C}_{head}$.
For the FFN block, which includes an up-projection $\W_{1} \in \mathbb{R}^{d \times d_{ff}}$ and a down-projection layer $\W_{2} \in \mathbb{R}^{d_{ff} \times d}$, where $d$ is the hidden dimension and $d_{ff}=4d$ is the number of intermediate neurons.
% typically equal to $4d$. 
Since it's time-cost to enumerate all $d_{ff}$ neurons, we divide them into $N$ groups and measure the MoPE of each group to obtain $\mathcal{C}_{neuron}$.
Then the insignificant heads and groups of neurons would be pruned, and we use knowledge distillation to transfer knowledge from the fixed teacher model to the final MoPE-CLIPw.
Second, we compress the MoPE-CLIPw in the depth direction.
We compute the MoPE for $L$ Transformer layers of MoPE-CLIPw and create the $\mathcal{C}_{layer}$.
With the assistance of $\mathcal{C}_{layer}$, we evaluate the priority of layers precisely and prune less important ones.
The final MoPE-CLIP is obtained by distilling from the fixed teacher model.

\vspace{-0.3cm}
\paragraph{Pre-training Stage.} 
We simultaneously compress the vision and text encoder of the large model to generate more general small models in the pre-training stage.
In addition to model capacity, training cost is another crucial challenge.
\lin{The width-first-then-depth strategy involves a two-stage retraining process, incurring high costs.
Moreover, the knowledge acquired in each retraining process expands as more image-text pairs are introduced during pre-training.} 
Therefore, we 
combine the \textbf{width-and-depth pruning} into a single stage, as depicted in Figure \ref{fig:pruning-process}(c).
In particular, we parallelly compute the MoPE metric for heads, groups of neurons, and layers of zero-shot CLIP's vision and text encoders.
After creating the cost tables, the pruning process is completed directly in several seconds.
Then we pre-train the pruned model on one small-scale image-text pre-training dataset (e.g., the CC3M dataset) and obtain the final MoPE-CLIP.
Our experiments in Section \ref{subsec:exp_pretraining} show 
that our MoPE-CLIP largely outperforms several efficient pre-training models \citep{lee2022uniclip,kim2023mcd,yang2023alip}, indicating that pruning of the large models provides a better initialization for pre-training.

\vspace{-0.3cm}
\paragraph{Pruning Efficiency.}

\lin{Calculating the MoPE metric for each module takes a few seconds, and computations for all modules can be parallelized.
Thus, the overall time of establishing cost tables is much less than a complete fine-tuning or pre-training process.}
Subsequently,
we can directly prune at different ratios to obtain a series of compact models.

% table for vision and text small model results
\begin{figure*}[!h]
\begin{minipage}{0.5\linewidth}
\resizebox{\linewidth}{!}{
\begin{tabular}{c|ccc|ccc}
\toprule
\multirow{2}{*}{Approach} & \multicolumn{3}{c|}{Vision Encoder} & \multicolumn{3}{c}{MSCOCO (5K test set)}           \\  % \cmidrule{2-4}  \cmidrule{5-7} 
                          & Width & Depth         & Parmas          & TR@1          & TR@5          & TR@10         \\ 
\midrule
Teacher Model             & 1024 & 24         & 304M            & 76.2         & 92.9         & 96.4         \\ 
% KD-B32 CLIP-VIT-B/32$^{\ddag}$                     
CLIP-VIT-B/32             & 768 & 12          & 88M             & 67.5         & 88.0         & 93.4         \\ 
% Substituted$_V$                
SE-CLIP$_V$         & 768 & 12          & 88M             & 56.1         & 81.0         & 89.1          \\ 
\rowcolor{orange!10} MoPE-CLIP$_V$          & 384       & 18          & 86M        & \textbf{69.7} & \textbf{90.4} & \textbf{95.0} \\
\bottomrule
\end{tabular}
}
\vspace{-3pt}
\tabcaption{ Image-to-text retrieval results of three small model architectures on MSCOCO dataset.
All models are trained with distillation.
% \lin{whether use $\ddag$ to denote the model (b c) is trained with KD?}
% The model with $\ddag$ denotes training with knowledge distillation.
}
\label{tab:vision_small_models} 
\end{minipage}
\hspace{15pt}
\begin{minipage}{0.5\linewidth}
\resizebox{\linewidth}{!}{
\begin{tabular}{c|ccc|ccc}
\toprule
\multirow{2}{*}{Approach} & \multicolumn{3}{c|}{Text Encoder} & \multicolumn{3}{c}{MSCOCO (5K test set)} \\ % \cmidrule{2-4}  \cmidrule{5-7} 
                          & Width     & Depth     & Parmas    & IR @1       & IR @5      & IR @10      \\ 
\midrule
Teacher Model             & 768       & 12        & 85M       & 58.8        & 82.8       & 89.5        \\
% KD-B32                    
CLIP-VIT-B/32             & 512       & 12        & 38M       & 49.4        & 75.8       & 84.7        \\
% SE-CLIP$_T$  Substituted$_T$              
SE-CLIP$_T$              & 512       & 12        & 38M       & 58.5        & 82.9       & 89.6        \\
\rowcolor{orange!10} MoPE-CLIP$_T$             & 384       & 12        & 42M       &\textbf{59.6} &\textbf{83.2} &\textbf{89.8}     \\ 
\bottomrule
\end{tabular}
}
\vspace{-3pt}
\tabcaption{ Text-to-image retrieval results of three small model architectures on MSCOCO dataset.
All models are trained with distillation.
}
\label{tab:text_small_models} 
\end{minipage}
\vspace{-0.8em}
\end{figure*}

% \subsection{Advanced Knowledge Distillation} \label{subsec:method_kd}

% \subsection{Distillation with Modality Knowledge} \label{subsec:method_kd}
\subsection{Distillation to MoPE-CLIP} \label{subsec:method_kd}

In contrast to previous distillation methods applied to ViT or BERT \citep{jiao2019tinybert, zhang2020ternarybert, yu2022unifiedvitcompress, yu2023vit_pruning_wu}, 
we design an advanced distillation loss that effectively transfers both cross-modal and uni-modal knowledge from 
% we design a more effective loss that transfers both cross-modal and uni-modal knowledge from 
large CLIP (teacher model) 
% teacher model 
to pruned MoPE-CLIP (student model) shown in Figure \ref{fig:pruning-process} (b).

\vspace{-0.3cm}
\paragraph{Cross-modal Knowledge.}
The CLIP model computes the cross-modal similarity matrix for retrieval and classification tasks. The teacher model exhibits a more closely aligned textual and visual embedding space, resulting in additional valuable knowledge within their similarity matrices.  To enhance the cross-modal capabilities of pruned models, we minimize the soft cross-entropy loss (SCE) between student similarity matrix $\mathbf{S}$ and teacher similarity matrix $\hat{\mathbf{S}}$, i.e.,
\begin{equation}
    \mathcal{L}_{sim} = \textrm{SCE} (\mathbf{S}, \hat{\mathbf{S}}).
    \vspace{-5pt}
\end{equation}

\paragraph{Uni-modal Knowledge.}
The teacher model possesses more substantial and superior vision or text encoders.
% respectively. 
Hence, it becomes crucial to transfer the knowledge embedded within these larger encoders to the student models.
Following \citep{jiao2019tinybert}, we utilize the mean squared error (MSE) loss to ensure that the student model's features $(\mathbf{F}_{v}, \mathbf{F}_{l})$ are as similar as possible to those of the teacher model $(\hat{\mathbf{F}}_{v}, \hat{\mathbf{F}}_{l})$:
\begin{equation}
    \mathcal{L}_{feat} = \frac{1}{2}\textrm{MSE} (\mathbf{F}_{v}, \hat{\mathbf{F}}_{v}) + \frac{1}{2}\textrm{MSE} (\mathbf{F}_{l}, \hat{\mathbf{F}}_{l}).
    \vspace{-5pt}
\end{equation}
Besides,  we also perform intermediate-layer distillation to transfer the hidden states knowledge (i.e., the output of each Transformer layer) $\mathbf{H}_v^m(m=1,2,...,M)$, $\mathbf{H}_l^k(k=1,2,...,K)$ from the teacher model to the student model. 
The depth-pruned student would mimic the preserved intermediate layers in the teacher model. The hidden loss is
\begin{equation}
    \mathcal{L}_{hidn}^{v}= \sum\nolimits_{m=1}^{M} \textrm{MSE} (\mathbf{H}_v^m, \hat{\mathbf{H}}_v^m),
    \vspace{-5pt}
\end{equation}
\begin{equation}
    \mathcal{L}_{hidn}^{l}= \sum\nolimits_{k=1}^{K} \textrm{MSE} (\mathbf{H}_l^k, \hat{\mathbf{H}}_l^k) ,
    \vspace{-5pt}
\end{equation}
\begin{equation}
    \mathcal{L}_{hidn}= \frac{1}{2} ( \mathcal{L}_{hidn}^{v} + \mathcal{L}_{hidn}^{l} ).
    \vspace{-5pt}
\end{equation}

\paragraph{Learning Objective.}
Combining the cross-modal knowledge and uni-modal knowledge, we further incorporate the contrastive loss ($\mathcal{L}_{itc}$).
Thus, the final training objective is 
% between the vision and text encoders of the small model to ensure alignment of their feature embeddings. 
% Therefore, the final training objective is 
\begin{equation} \label{eq:dis_loss}
    \mathcal{L} = \mathcal{L}_{itc} + \alpha  \mathcal{L}_{sim} + \beta \mathcal{L}_{feat} + \gamma  \mathcal{L}_{hidn}
    \vspace{-5pt}
\end{equation}
By default, we do not tune and set $(\alpha, \beta, \gamma)=(1,10^3,1)$ to ensure a balanced magnitude of these losses.

% balance these losses to a similar magnitude, ensuring a more robust optimization process.

\section{Experiments}

% % =================== Version 1 ======================
% % Divide the two-stage into two sections and introduce them independently

\subsection{Fine-tuning Stage Compression} \label{subsec:exp_finetuning}

% table for vision pruning analysis
\begin{table}[h]
\resizebox{\linewidth}{!}{
\begin{tabular}{c|ccc|ccc}
\toprule
\multirow{2}{*}{Pruning}    & \multicolumn{3}{c|}{Vision Encoder} & \multicolumn{3}{c}{MSCOCO (5K test set)} \\ % \cmidrule{2-4}  \cmidrule{5-7}
                            & Width      & Depth     & Parmas     & TR @1       & TR @5      & TR @10      \\ \midrule
Teacher Model               & 1024       & 24        & 304M       & 76.2        & 92.9       & 96.4        \\ \midrule
\multirow{2}{*}{MagnCLIP$_V$ \citep{han2015learning}}  
                            & 512        & 24        & 153M       & 71.2        & 90.8       & 95.2     \\
                            & 384        & 24        & 115M       & 64.2        & 86.6       & 92.8        \\ \midrule
\multirow{3}{*}{DynaCLIP$_V$ \citep{hou2020dynabert}}  
                            & 512        & 24        & 153M       & 73.9        & 92.0       & 96.0     \\
                            & 384        & 24        & 115M       & 70.3        & 90.0       & 94.9        \\
                            & 384        & 18        & 86M        & 67.6        & 88.7       & 94.1        \\ \midrule
% UPop-CLIP\citep{shi2023upop}   & N/A        & N/A        & *         & 56.1                                     & 82.4       & 90.2        \\ \midrule
\multirow{2}{*}{UPop-CLIP \citep{shi2023upop}}      
                            & N/A        & N/A       & 474M$^{\ddag}$ 
                            & 70.8        & 90.8       & 95.2        \\                        
                            & N/A        & N/A        & 280M$^{\ddag}$         & 56.1         
                            & 82.4       & 90.2        \\ \midrule
% \rowcolor{orange!10}
\multirow{3}{*}{MoPE-CLIP$_V$} & 512        & 24        & 153M       & \textbf{74.7} & \textbf{92.2} & \textbf{96.4} \\
                            & 384        & 24        & 115M       & \textbf{72.1} & \textbf{91.5} & \textbf{95.7} \\
                            & 384        & 18        & 86M        & \textbf{69.7} & \textbf{90.4} & \textbf{95.0} \\
\bottomrule
\end{tabular}
}
\vspace{-3pt}
\caption{Image-to-text retrieval results of different pruning methods on the MSCOCO dataset with several pruning ratios.
The Params labeled as $^{\ddag}$ denote the parameters of the entire model.
}
\label{tab:vision_pruning} 
% \vspace{-1.em}
\end{table}

% table for text compress
\begin{table}[h]
\resizebox{\linewidth}{!}{
\begin{tabular}{c|ccc|ccc}
\toprule
\multirow{2}{*}{Pruning}     & \multicolumn{3}{c|}{Text Encoder} & \multicolumn{3}{c}{MSCOCO (5K test set)}        \\
                              & Width     & Depth     & Params    & IR @1         & IR @5         & IR @10        \\ \midrule
Teacher Model                 & 768       & 12        & 85M       & 58.8          & 82.8          & 89.5          \\ \midrule
\multirow{2}{*}{MagnCLIP$_T$ \citep{han2015learning}}  & 384       & 12        & 42M       & 59.2          & 82.9          & 89.1          \\
                              & 192       & 12        & 21M       & 56.6          & 81.9          & 89.2          \\ \midrule
\multirow{2}{*}{DynaCLIP$_T$ \citep{hou2020dynabert}}  & 384       & 12        & 42M       & 59.3          & 83.0         & 89.7        \\
                              & 192       & 12        & 21M       & 57.3          & 82.3          & 89.4          \\ \midrule
\multirow{2}{*}{MoPE-CLIP$_T$}  & 384       & 12        & 42M       & \textbf{59.6} & \textbf{83.2} & \textbf{89.8} \\
                              & 192       & 12        & 21M       & \textbf{58.0} & \textbf{82.6} & \textbf{89.8} \\ 
\bottomrule
\end{tabular}
}
\vspace{-3pt}
\caption{Text-to-image retrieval results of different pruning methods on the MSCOCO dataset with two pruning ratios.
% Pruning is applied to the width direction. 
% The ''Params'' exclude the parameters of the Embedding Layer.
}
\label{tab:text_compress} 
% \vspace{-1.em}
\vspace{-0.5cm}
\end{table}

\paragraph{Experimental Settings.}
During the fine-tuning stage, we compress fine-tuned CLIP-ViT-L/14 (FT-L14) to create task-specific models.
We select cross-modal retrieval as our downstream tasks and evaluate the compressed model on the MSCOCO \citep{lin2014mscoco} and Flickr30K \citep{plummer2015Flickr30K} datasets.
Due to limited space, the results on Flickr30K are in Appendix \ref{subsec:appx_flickr}.

\vspace{-0.2cm}
\paragraph{Implementation Details.} 
We apply the width-first-then-depth pruning on the vision or text encoder of FT-L14 to obtain MoPE-CLIP$_V$ and MoPE-CLIP$_T$. 
The MoPE metric is computed by TR Mean and IR Mean, respectively.
In particular, since individually processing all neurons is time-consuming, we first rewire all FFN neurons according to loss gradient like \citep{hou2020dynabert}, then divide them into groups for acceleration.
% to accelerate the computation.
Knowledge distillation is added to enhance the performance of fine-tuned CLIP-ViT-B/32 and SE-CLIP. 
% in Figure \ref{fig:small-models}.

% zero-shot retrieval evaluation
\begin{table*}[!t]
\begin{center}
\resizebox{\linewidth}{!}{
\begin{tabular}{l|cc|cc|c|cccccc|cccccc}
\toprule
\multirow{2}{*}{Method} & \multicolumn{2}{c|}{Vision Enocder} & \multicolumn{2}{c|}{Text Encoder} & Params(M) & \multicolumn{6}{c|}{MSCOCO (5K test set)}                                                     & \multicolumn{6}{c}{Flickr30K (1K   test set)}                                        \\ % \cmidrule{2-4}  \cmidrule{5-7}
                          & Width            & Depth            & Width           & Depth           & Vision + Text                           & TR @1         & TR @5         & TR @10        & IR @1         & IR @5         & IR @10        & TR @1         & TR @5         & TR @10        & IR @1         & IR @5         & IR @10        \\ 
\midrule
\color{gray}{\textit{Pre-trained on WIT-400M}} \\
CLIP-ViT-L/14 \citep{radford2021clip}            & 1024             & 24               & 768             & 12              & 304 + 85                   & 56.3          & 79.4          & 86.6          & 36.5          & 61.1          & 71.2          & 85.2          & 97.5          & 99.1          & 64.9          & 87.3          & 92.2          \\
CLIP-ViT-B/32 \citep{radford2021clip}            & 768              & 12               & 512             & 12              & 88 + 38                 & 50.1          & 75.0          & 83.5          & 30.5          & 56.0          & 66.9          & 78.8          & 94.9          & 98.2          & 58.8          & 93.6          & 90.2          \\ 
\midrule
\color{gray}{\textit{Pre-trained on CC3M}} \\
EfficientVLM \citep{wang2022efficientvlm}        & 1024             & 12               & 768             & 6               & 152 + 42             & 46.6          & 71.7          & 81.3          & 35.9          & 61.6          & 71.8          & 78.8          & 94.9          & 98.2          & 58.8          & 93.6          & 90.2    \\
TinyCLIP \citep{wu2023tinyclip}              & 512             & 24               & 768             & 6               & 152 + 42                & 52.7          & 76.5          & 84.8          & 36.6          & 63.0          & 73.6          & 80.5          & 96.3          & 98.5          & 66.3          & 89.1          & 93.7    \\

\rowcolor{orange!10} 
MoPE-CLIP$_{large}$     & 512             & 24               & 384             & 12              & 152 + 42                 & \textbf{58.0} & \textbf{81.6} & \textbf{88.5} & \textbf{40.6} & \textbf{66.0} & \textbf{75.5} & \textbf{86.5} & \textbf{97.7} & \textbf{99.0} & \textbf{69.8} & \textbf{90.6} & \textbf{95.3} \\
% \cdashlinelr{1-18}

DynaCLIP$_{base}$ \citep{hou2020dynabert}        & 384              & 18               & 384             & 12              & 86 + 42                    & 51.3          & 75.5          & 84.6          & 35.8          & 61.8          & 72.6          & 79.8          & 96.1          & 98.2          & 64.6          & 87.8          & 93.1          \\
DynaCLIP$_{small}$ \citep{hou2020dynabert}        & 384              & 18               & 192             & 12              & 86 + 21                   & 46.7          & 72.7          & 92.2          & 33.2          & 59.5          & 70.3          & 75.9          & 94.6          & 98.3          & 60.9          & 86.1          & 91.9          \\
\rowcolor{orange!10} MoPE-CLIP$_{base}$        & 384              & 18               & 384             & 12              & 86 + 42                    & \textbf{52.8} & \textbf{78.1} & \textbf{86.0} & \textbf{37.3} & \textbf{63.5} & \textbf{73.6} & \textbf{82.8} & \textbf{97.1} & \textbf{98.8} & \textbf{66.7} & \textbf{88.7} & \textbf{94.1} \\
\rowcolor{orange!10} MoPE-CLIP$_{small}$       & 384              & 18               & 192             & 12              & 86 + 21                    &  \textbf{50.3} & \textbf{75.9} & \textbf{84.8} & \textbf{35.6} & \textbf{61.7} & \textbf{72.2} & \textbf{80.2} & \textbf{95.6} & \textbf{98.5} & \textbf{64.7} & \textbf{87.8} & \textbf{93.0}  \\ 

\midrule
\color{gray}{\textit{Pre-trained on YFCC15M}} \\
CLIP-ViT-B/32$^{\dag}$ \citep{radford2021clip}    & 768              & 12               & 512             & 12              & 88 + 38                    & 20.8          & 43.9          & 55.7          & 13.0           & 31.7          & 42.7          & 34.9          & 63.9          & 75.9          & 23.4          & 47.2          & 58.9          \\
SLIP-ViT-B/32$^{\dag}$ \citep{mu2022slip}   & 768                   & 12               & 512             & 12              & 88 + 38                    & 27.7          & 52.6          & 63.9          & 18.2          & 39.2          & 51.0            & 47.8          & 76.5          & 85.9          & 32.3          & 58.7          & 68.8          \\
DeCLIP-ViT-B/32$^{\dag}$ \citep{li2021declip} & 768                 & 12               & 512             & 12              & 88 + 38                    & 28.3          & 53.2          & 64.5          & 18.4          & 39.6          & 51.4          & 51.4          & 80.2          & 88.9          & 34.3          & 60.3          & 70.7          \\
UniCLIP-ViT-B/32$^{\dag}$ \citep{lee2022uniclip} & 768              & 12               & 512             & 12              & 88 + 38                    & 32.0          & 57.7          & 69.2          & 20.2          & 43.2          & 54.4          & 52.3          & 81.6          & 89.0          & 34.8          & 62.0            & 72.0            \\
MCD-ViT-B/32$^{\dag}$ \citep{kim2023mcd}     & 768                  & 12               & 512             & 12              & 88 + 38                    & 32.2          & 58.7          & 71.2          & 20.7          & 43.5          & 55.3          & 57.6          & 82.6          & 91.1          & 36.4          & 64.8          & 74.1          \\
ALIP-ViT-B/32$^{\dag}$ \citep{yang2023alip}    & 768                & 12               & 512             & 12              & 88 + 38                    & 46.8          & 72.4          & 81.8          & 29.3          & 54.4          & 65.4          & 70.5          & 91.9          & 95.7          & 48.9          & 75.1          & 82.9          \\
\rowcolor{orange!10} MoPE-CLIP$_{base}$        & 384                & 18               & 384             & 12              & 86 + 42                    & \textbf{55.6} & \textbf{78.6} & \textbf{86.1} & \textbf{37.1} & \textbf{63.1} & \textbf{73.5} & \textbf{86.1} & \textbf{97.9} & \textbf{99.6} & \textbf{66.4} & \textbf{89.2} & \textbf{94.2} \\ 
\bottomrule
\end{tabular}
}
\end{center}
\vspace{-10pt}
\caption{Zero-shot image-text retrieval results on MSCOCO and Flickr30K datasets.
Our MoPE-CLIP$_{base}$ pre-trained on CC3M datasets outperforms the CLIP-ViT-B/32 pre-trained on WIT-400M on all the metrics.
$^{\dag}$ denotes the results are reported from \citep{lee2022uniclip,kim2023mcd,yang2023alip}.
}
\vspace{-0.7em}
\label{tab:pretrain-zs}
\end{table*}

\vspace{-0.2cm}
\paragraph{Further evaluation of three small model architectures.}
Table~\ref{tab:vision_small_models} and Table \ref{tab:text_small_models} present the image-to-text retrieval and text-to-image retrieval performance of three architectures, respectively.
With similar parameters, our MoPE-CLIP$_V$ performs best and surpasses CLIP-ViT-B/32 by 2.2\% TR@1 and SE-CLIP$_{V}$ by 13.6\% TR@1.
MoPE-CLIP$_{T}$ at 2x compression ratio also outperforms SE-CLIP$_{T}$ and CLIP-ViT-B/32.
These results indicate that compared to the \textbf{pretrained} small models and \textbf{substituted} encoder models, MoPE-CLIP$_V$ and MoPE-CLIP$_T$ provide better small CLIP models while maintaining flexibility.
Additionally, we observe that the knowledge distillation process actually improves the TR@1 of CLIP-ViT-B/32 and SE-CLIP$_V$ in Figure \ref{fig:small-models}.
This demonstrates the effectiveness of the teacher's knowledge, but the architectural difference between ViT-L14 and ViT-B32 limits the final performance.

\vspace{-0.3cm}
\paragraph{Comparison with Other Pruning Methods.}
We compare our models with the state-of-the-art VLP compression method UPop \citep{shi2023upop}.
We also extend the uni-modal pruning methods on CLIP architecture, including
the dynamic pruning method DynaBERT \citep{hou2020dynabert} and magnitude-based pruning \citep{han2015learning}.
Notably, distillation is applied to DynaCLIP and MagnCLIP, except for Upop whose result is from the original paper.
As seen in Table \ref{tab:vision_pruning}, MoPE-CLIP$_V$ performs significantly better than other DynaCLIP$_V$ and MagnCLIP$_V$ at the same depth and width, especially the TR@1.
Compared with UPop, our MoPE-CLIP$_{V}$ with 153M vision encoder termed an entire model of 234M parameters largely surpasses the UPop-CLIP with 474M parameters on all metrics.
In addition, Table \ref{tab:text_compress} shows that even at the 4x compression ratio, our MoPE-CLIP$_{T}$ still maintains high performance
on the text-to-image retrieval task, 
with only a 0.8\% drop in IR@1 compared to the teacher model. 
We report performance under more pruning ratios and compare Upop with KD in Appendix \ref{subsec:appx_comparison}.
We analyze the difference of preserved heads between MoPE-CLIP$_V$ and DynaCLIP$_V$ in 
% \textit{supplementary material}, 
Appendix \ref{subsec:appx_analysis_pruing},
which further demonstrates the accurate assessment of MoPE metrics.

% ###########################Pre-training stage compression#########################

\subsection{Pre-training Stage Compression}\label{subsec:exp_pretraining}

% zero-shot classification  performance
\begin{table*}[!ht]
\begin{center}
\resizebox{\linewidth}{!}{
% \begin{tabular}{l|cc|ccccccccccc|c}
\begin{tabular}{lcccccccccccccc}
\toprule
Method
& \shortstack{Pre-training \\ dataset}
& \shortstack{Training \\ epochs}
& \rotatebox{90}{CIFAR10}              
& \rotatebox{90}{CIFAR100}             
& \rotatebox{90}{Caltech101}           
& \rotatebox{90}{Flowers}              
& \rotatebox{90}{Pets}           
& \rotatebox{90}{DTD}                  
& \rotatebox{90}{Cars}                 
& \rotatebox{90}{Aircraft}     
& \rotatebox{90}{SUN397}               
& \rotatebox{90}{Food101}
& \rotatebox{90}{ImageNet}             
& \rotatebox{90}{\textbf{Average}} 
\\ 
\midrule
CLIP-ViT-B/32$^{\dag}$ \citep{radford2021clip}      & YFCC15M       & 50     & 62.3    & 33.6     & 55.4       & 6.3     & 19.4 & 16.9 & 2.1  & 1.4      & 40.2   & 33.7    & 31.3     & 27.5 \\
SLIP-ViT-B/32$^{\dag}$ \citep{mu2022slip}           & YFCC15M       & 50     & 72.2    & 45.3     & 65.9       & 6.8     & 28.3 & 21.8 & 2.9  & 1.9      & 45.1   & 44.7    & 38.3     & 33.9 \\
DeCLIP-ViT-B/32$^{\dag}$ \citep{li2021declip}       & YFCC15M       & 50     & 72.1    & 39.7     & 70.1       & 7.1     & 30.2 & 24.2 & 3.9  & 2.5      & 41.6   & 46.9    & 39.2     & 34.3 \\
UniCLIP-ViT-B/32$^{\dag}$ \citep{lee2022uniclip}    & YFCC15M       & 50     & 78.6    & 47.2     & 73.0       & 8.1     & 32.5 & 23.3 & 3.4  & 2.8      & 50.4   & 48.7    & 41.2     & 37.2 \\
MCD-ViT-B/32$^{\dag}$ \citep{kim2023mcd}            & YFCC15M       & 32     & 80.3    & 49.6     & 73.2       & 7.9     & 40.0 & 30.5 & 3.4  & 3.0      & 55.3   & 54.0    & 44.7     & 40.2 \\
ALIP-ViT-B/32$^{\dag}$ \citep{yang2023alip}         & YFCC15M       & 32     & 83.8    & 51.9     & 74.1       & 54.8    & 30.7 & 23.2 & 5.4  & 2.7      & 47.8   & 45.4    & 40.3     & 41.8 \\ 
\rowcolor{orange!10}
MoPE-CLIP$_{base}$ & YFCC15M       & 20     & \textbf{91.5} & \textbf{68.1} & \textbf{85.5} & \textbf{66.8} & 69.3   & \textbf{46.6} & 16.6    & 6.0    & 61.2   & \textbf{74.6} & \textbf{60.7} & 58.8   \\ 
\midrule
\rowcolor{orange!10}
MoPE-CLIP$_{base}$ & CC3M          & 20     & 86.8    & 61.7     & 79.0       & 30.1    & 42.0          & 38.5  & 5.6           & 1.7          & 57.1          & 38.6     & 44.5     & 44.2  \\
\rowcolor{orange!10}
MoPE-CLIP$_{base}$ & CC12M         & 20     & 91.2    & 67.3     & 85.0       & 45.0    & \textbf{80.0} & 41.1  & \textbf{47.7} & \textbf{7.2} & \textbf{62.2} & 70.6     & \textbf{60.7}  & \textbf{59.8} \\
\bottomrule
\end{tabular}
}
\end{center}
\vspace{-10pt}
\caption{ Top-1 accuracy(\%) of zero-shot image classification on 11 downstream datasets.
Our MoPE-CLIP$_{base}$ largely surpasses other state-of-the-art efficient pre-training methods using fewer training epochs.
$^{\dag}$ denotes the results are reported from \citep{lee2022uniclip,kim2023mcd,yang2023alip}.
% Comparison of MoPE-CLIP with other efficient pre-training methods on image classification tasks. 
% $^{\dag}$ denotes models are pre-trained on the YFCC15M dataset and the results are reported from \citep{lee2022uniclip}.
}
\vspace{-0.9em}
\label{tab:pretrain-classification}
\end{table*}

\paragraph{Experimental Setting}

During the pre-training stage, we compress the zero-shot CLIP-ViT-L/14 (ZS-14) model \citep{radford2021clip} to obtain compact general models. 
% Subsequently, we pre-train various pruned models on a small-scale pre-training dataset, CC3M \citep{sharma2018cc3m}.
Subsequently, we pre-train our MoPE-CLIP and various baselines on a small-scale pre-training dataset, CC3M \citep{sharma2018cc3m}.
To further assess the capabilities of our MoPE-CLIP model, we scale up training using larger datasets, including CC12M \citep{changpinyo2021cc12m} and YFCC15M \citep{li2021declip}.
% Cross-modal retrieval and image classification are selected as our evaluation tasks.
% Our evaluation focuses on cross-modal retrieval and image classification tasks.
In addition, we evaluate our pruning method on OpenCLIP ViT-B/16 and report results in Appendix \ref{subsec:appx_openclip}.

\vspace{-0.2cm}
% \paragraph{Model Architectures.}
% Implementation details --- all our models
\paragraph{Implementation Details}
% During the pre-training stage, 
We simultaneously prune both vision and text encoders of ZS-L14.  
For the vision encoder, we adopt width-and-depth pruning and compress the encoder to 86M parameters, which is similar to CLIP-ViT-B/32. 
For the text encoder, we compress it in the width direction at two pruning ratios, resulting in MoPE-CLIP$_{base}$ and MoPE-CLIP$_{small}$. 
% To ensure a fair comparison with prior state-of-the-art VLP compression methods \citep{wang2022efficientvlm, wu2023tinyclip}, we also prune both the vision and text encoders to half-width, producing MoPE-CLIP$_{large}$.
We also prune both the vision and text encoders to half-width, producing MoPE-CLIP$_{large}$.
The module's importance is evaluated on the MSCOCO validation dataset and the Recall Mean serves as the MoPE metric.
More details are left in Appendix \ref{sec:appx_implementation}.

\vspace{-0.2cm}
\paragraph{Zero-shot Image-text Retrieval.}
Table \ref{tab:pretrain-zs} shows the zero-shot retrieval results on MSCOCO and Flickr30K datasets. 
% Both MoPE-CLIP$_{base}$ and DynaCLIP$_{base}$ consistently surpass the CLIP-ViT-B32 in all Recall metrics, especially for the text-to-image retrieval task.
MoPE-CLIP$_{base}$ consistently surpasses the CLIP-ViT-B32 in all Recall metrics.
MoPE-CLIP$_{small}$ maintains competitive results and outperforms the DynaCLIP$_{small}$ with a clear margin.
In addition, when compared with previous efficient pre-training methods,
% \citep{li2021declip,lee2022uniclip,kim2023mcd,yang2023alip},
MoPE-CLIP$_{base}$ pre-trained on CC3M achieves 52.8\% TR@1 and 37.3\% IR@1 on the MSCOCO dataset, which is 6.0\% and 8.0\% higher than ALIP \citep{yang2023alip} pre-trained on YFCC15M.
The improvement is mainly attributed to the pruned large model providing a better initialization for pre-training vision-language models.

\vspace{-0.2cm}
\paragraph{Zero-shot Classification.}
\lin{
We adopt the Recall Mean on the MSCOCO validation dataset as the MoPE metric, which reflects the module influence of multi-modal tasks. To demonstrate the robustness of Recall Mean on uni-modal tasks, we further compare our MoPE-CLIP$_{base}$ with other efficient pre-training methods on zero-shot image classification tasks. 
}
SLIP \citep{mu2022slip}, DeCLIP \citep{li2021declip}, and UniCLIP \citep{lee2022uniclip} incorporate fine-grained supervision to reduce the data requirement. 
ALIP \citep{yang2023alip} and MCD \citep{kim2023mcd} propose new frameworks to reduce the noise and misalignments in image-text pairs.
We utilize the same prompt templates following CLIP \citep{han2015learning}.
Table \ref{tab:pretrain-classification} presents the results on 11 widely used benchmarks.
Our MoPE-CLIP$_{base}$ pre-trained on the YFCC15M dataset significantly surpasses previous methods and creates new state-of-the-art results, indicating the effectiveness of MoPE-CLIP towards classification tasks.

\vspace{-0.3cm}
% Need to discuss and further revise
\paragraph{Comparison with VLP Compression Methods.}

\begin{figure}[t]
    \centering
    \includegraphics[width=0.95\linewidth]{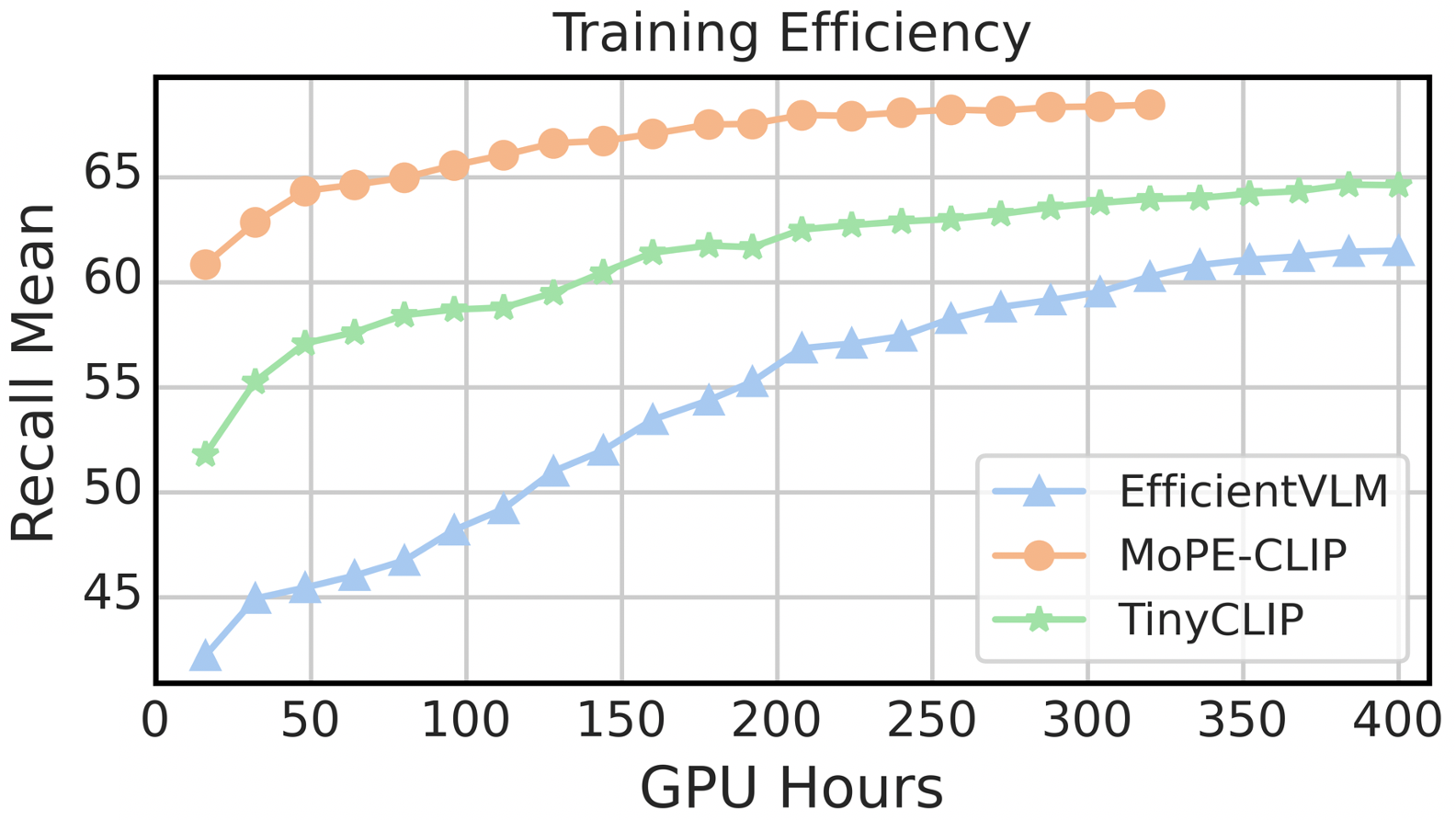}
    \vspace{-5pt}
    \caption{Comparsion of training efficiency. EfficientVLM and TinyCLIP are trained for 25 epochs, while MoPE-CLIP$_{large}$ is trained for 20 epochs. All models are compressed from CLIP-ViT-L/14 at a 2x compression ratio and trained on the CC3M dataset. 
    }
    \label{fig:training_efficiency}
    % \vspace{-1em}
\end{figure}

We employ state-of-the-art vision-language compression methods, EfficientVLM \citep{wang2022efficientvlm} and TinyCLIP \citep{wu2023tinyclip}, to compress the zero-shot CLIP-ViT-L/14 model. 
These compressed models are then pre-trained on the CC3M dataset using the respective loss functions.
The zero-shot retrieval results, as presented in Table \ref{tab:pretrain-zs}, unequivocally illustrate that our MoPE-CLIP$_{large}$ performs the best.
Notably, even our MoPE-CLIP$_{base}$ with a reduction to 66M parameters still surpasses the TinyCLIP and EfficientVLM.
We further compare the training process of 
these models
in Figure \ref{fig:training_efficiency}.
Our models achieve competitive results in less training time, highlighting the significant contribution of our MoPE metric in preserving crucial modules. 
In contrast to TinyCLIP, which focuses on cross-modal affinity, and EfficientVLM, which emphasizes uni-modal knowledge transfer, our approach combines cross-modal and uni-modal distillation, proving more effective in enhancing pruned model capacity.

\vspace{-0.3cm}
\paragraph{Inference Speedup.}
We measure the latency using PyTorch inference mode with the batch size of 64 in Table~\ref{tab:inference_speed}, where MoPE-CLIP shows a significant speedup.

\vspace{-0.3cm}
\paragraph{Training Cost.}
The entire training process for MoPE-CLIP on the CC3M dataset only requires 40 hours on 8x NVIDIA V100 GPUs, suggesting that pruning provides a superior solution for achieving general compact VLP models with flexibility and minimal training cost.

\begin{table}[t]
\centering
\resizebox{1.\linewidth}{!}{
\begin{tabular}{c|cccc}
\toprule
Models        & CLIP-ViT-L/14 & MoPE$_{large}$ & MoPE$_{base}$ & MoPE$_{small}$ \\ 
\midrule
Params (M)     & 390   & 194 ${\color{red}\downarrow 50\%}$  & 128 ${\color{red}\downarrow 67\%}$  & 107 ${\color{red}\downarrow 73\%}$           \\
Latency (ms)     & 141.96   & 79.00 ${\color{red}\downarrow 44\%}$   & 58.73 ${\color{red}\downarrow 59\%}$  & 49.48 ${\color{red}\downarrow 65\%}$          \\ 
\bottomrule
\end{tabular}
}
% \vspace{-5pt}
\caption{Nvidia V100 GPU latency (ms) on MSCOCO test sets.}
\label{tab:inference_speed} 
\vspace{-1.2em}
\end{table}

% table for layer selection strategy
\begin{table}[ht]
\resizebox{\linewidth}{!}{
\begin{tabular}{c|c|ccc}
\toprule
\multirow{2}{*}{Setting}                                                                                      & \multirow{2}{*}{Method} & \multicolumn{3}{c}{COCO test set}                         \\ \cmidrule{3-5} 
                                                                                                              &                         & TR@1          & TR@5          & \multicolumn{1}{l}{TR@10} \\ \midrule
\multirow{5}{*}{\begin{tabular}[c]{@{}c@{}}Prune 3 layers \\ for \\ 0.375width \\ MWPE-CLIP$_V$\end{tabular}} & Top Layers              & 70.1          & 90.2          & 95.4                      \\
                                                                                                              & Bottom Layers           & 70.8          & 90.4          & 95.1                      \\
                                                                                                              & Every Other             & 69.2          & 90.0          & 94.9                      \\
                                                                                                              & Loss Gradient           & 70.4          & 90.6          & 94.9                      \\
                                                                                                              
&\cellcolor{orange!10}\textbf{MWPE metric}    & \cellcolor{orange!10}\textbf{72.2} &\cellcolor{orange!10} \textbf{91.2} & \cellcolor{orange!10}\textbf{95.5}             \\ \midrule
\multirow{5}{*}{\begin{tabular}[c]{@{}c@{}}Prune 6 layers \\ for \\ 0.375width\\ MWPE-CLIP$_V$\end{tabular}}  & Top Layers              & 57.6          & 81.7          & 88.6                      \\
                                                                                                              & Bottom Layers           & 63.9          & 88.0          & 93.5                      \\
                                                                                                              & Every Other             & 66.6          & 88.9          & 93.6                      \\
                                                                                                              & Loss Gradient           & 66.3          & 87.8          & 94.0                      \\
                                                                                                              
&\cellcolor{orange!10}\textbf{MWPE metric}    &\cellcolor{orange!10}\textbf{69.7} &\cellcolor{orange!10}\textbf{90.4} & \cellcolor{orange!10}\textbf{95.0}             \\ \bottomrule
\end{tabular}
}
\caption{Ablation study of Layer Selection strategies.
}
\label{tab:Layer_selection} 
\vspace{-0.5em}
\end{table}

\begin{table}[ht]
\centering
\resizebox{0.94\linewidth}{!}{
\begin{tabular}{ll|cccc}
\toprule
\multicolumn{2}{l|}{\multirow{2}{*}{MoPE-CLIP$_{V}$}} & \multicolumn{4}{c}{MSCOCO}                       \\
\multicolumn{2}{l|}{}                                 & TR@1 & TR@5 & TR@10 & \multicolumn{1}{l}{TRMean} \\ 
\midrule
\multicolumn{2}{l|}{Depth-first-then-width}           & 64.0 & 87.0 & 92.4  & 81.1                       \\
\multicolumn{2}{l|}{Width-and-depth}                  & 61.4 & 84.8 & 90.9  & 79.0                       \\
\multicolumn{2}{l|}{Width-first-then-depth}           & \textbf{69.7} & \textbf{90.4} & \textbf{95.0} & \textbf{85.0}                        \\ 
\bottomrule
\end{tabular}
}
\caption{ Ablation study in pruning 86M MoPE-CLIP$_V$.}
\label{tab:ablation_framework} 
\vspace{-0.5em}
\end{table}

% table for kd
\begin{table}[!t]
\centering
\resizebox{0.94\linewidth}{!}{
\begin{tabular}{l|cccc}
\toprule
\multirow{2}{*}{Training Loss} & \multicolumn{4}{c}{MSCOCO}                                             \\ % \cline{2-5} 
                                   & TR@1           & TR@5           & TR@10          & \multicolumn{1}{l}{TRMean} \\ 
\midrule
MoPE-CLIP$_V$                      & \textbf{69.7} & \textbf{90.4} & \textbf{95.0} & \textbf{85.0}              \\ 
\midrule
w/o $\mathcal{L}_{sim}$                            & 68.8          & 89.8          & 94.7          & 84.4                       \\
w/o $\mathcal{L}_{feat}$                             & 69.4          & 89.4          & 94.7          & 84.5                       \\
w/o $\mathcal{L}_{hidn}$                           & 67.7          & 89.1          & 94.0          & 83.6                       \\
w/o $Distillation$                   & 60.8          & 84.9          & 91.5          & 79.0                       \\
\bottomrule
\end{tabular}
}
\caption{ Ablation study of knowledge distillation.}
\label{tab:kd_loss} 
\vspace{-1.5em}
\end{table}

% ###########################Ablation Study#########################
\subsection{Ablation Study}
\label{subsec:ablation}

\paragraph{Effects of MoPE Metric.}
To further demonstrate the effectiveness of our MoPE metric, we conduct an ablation study for depth pruning. 
% Leveraging insights from prior works on layer reduction strategies in BERT \citep{fan2019reducingbertdepth, sajjad2020poormanbert}, we apply these techniques to the 0.375-width MoPE-CLIP$_V$ models. 
We apply four commonly used layer reduction strategies in BERT \citep{fan2019reducingbertdepth, sajjad2020poormanbert} to the 0.375-width MoPE-CLIP$_V$ models, including
% Specifically, we explore four strategies: 
(i) removal of either the bottom or top layers, (ii) the ``Every Other" strategy, and (iii) a Gradient-based approach that gauges layer importance by analyzing gradients with respect to all heads and neurons within a layer.
As presented in Table~\ref{tab:Layer_selection}, our MoPE metric outperforms other strategies with a clear margin.
Notably, the Every Other strategy falls behind when pruning three layers.
% indicating that this widely used strategy in the BERT model may not be the best choice. 
\lin{We assume that simply reducing every other layer in Transformer encoder may not influence the model capacity of uni-model tasks as proven in \citep{fan2019reducingbertdepth}.
However, the unavailability of the other encoder results in a performance drop in cross-modal tasks.
% but the inaccessibility of another encoder leads to the performance drop on cross-modal tasks.
}
These findings indicate the importance of selecting an appropriate strategy for layer reduction in CLIP models and our MoPE metric provides a straightforward yet valuable approach.

\begin{figure}[!t]
    \centering
    \includegraphics[width=0.98\linewidth]{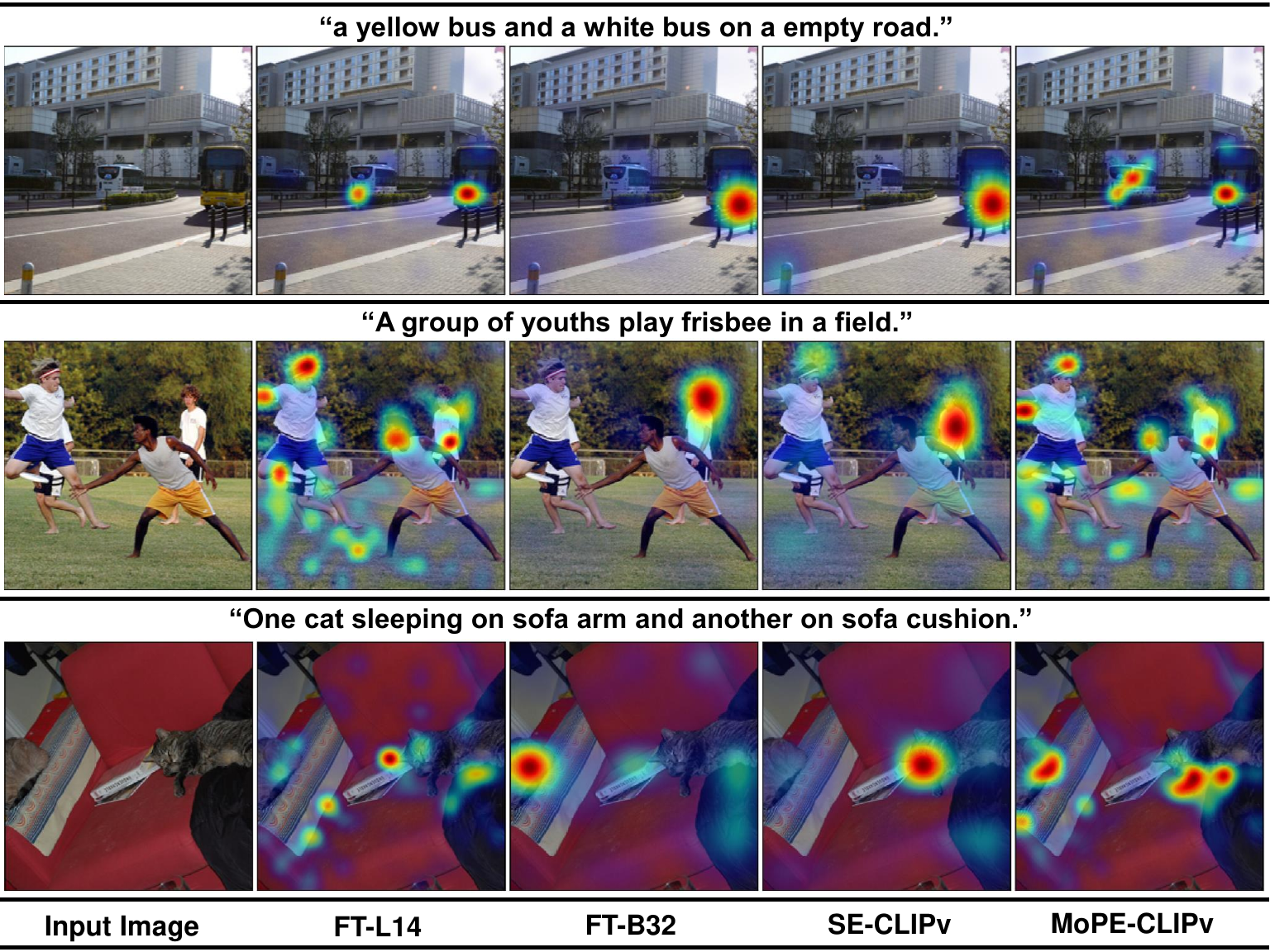}
    \caption{Grad-CAM visualization on the self-attention maps in the last layer of vision encoder for different models.}
    \label{fig:cam_diff_models}
    \vspace{-1.5em}
\end{figure}

\vspace{-0.3cm}
\paragraph{Effects of Pruning Framework.}
During the fine-tuning stage, we further explore the other two strategies, including pruning in a ``depth-first" manner followed by ``width pruning," as well as simultaneous ``width-and-depth" pruning. 
As depicted in Table \ref{tab:ablation_framework}, the ``width-first-then-depth" strategy yields the best performance, while the ``depth-first-then-width" and ``width-and-depth" strategies fall behind. 
This discrepancy may be attributed to the sequential computation of hidden states across different layers, making it challenging to accurately evaluate the importance of heads or neurons in layer-reduced models, as discussed in \citep{hou2020dynabert}. 
Furthermore, the fine-tuning dataset may not be sufficiently large to fully restore the model's capacity.
Therefore, during the fine-tuning stage, the ``width-first-then-depth" strategy stands out as the optimal choice for creating more competitive smaller models. 
In contrast, during the pre-training stage, adopting the ``width-and-depth" pruning strategy is more convenient and efficient, with performance recovery facilitated by a large corpus of image-text pairs.

\vspace{-0.2cm}
\paragraph{Effects of Knowledge Distillation.}
\label{subsec:ablation_kd}

We conduct an ablation study of our distillation objectives designed in Section \ref{subsec:method_kd}.
We investigate the learning process on MoPE-CLIP$_V$ and the results in Table \ref{tab:kd_loss} show the effectiveness of all our distillation loss.
We observe that all distilled models outperform models without distillation by a clear margin, demonstrating the importance of both cross-modal and uni-modal knowledge.
Importantly, the TR@1 of the ``w/o $\mathcal{L}_{hidn}$'' model drops significantly from 69.7\% to 67.7\%, which indicates the intermediate layer knowledge in the teacher model is crucial for retraining the MoPE-CLIP model.
However, the discrepancy in patch numbers between the 
ViT-B/32
and ViT-L/14 leads to the failure of hidden distillation applied to CLIP-ViT-B32 and SE-CLIP$_{V}$. 
Consequently, the effectiveness of knowledge distillation is largely diminished for pre-trained small models and substituted encoder models. 
In contrast, MoPE-CLIP shares a similar architecture with the teacher model, allowing it to acquire more knowledge.

\vspace{-0.2cm}
\paragraph{Visualization.}

To better understand the architecture influence on the retrieval task, we utilize Grad-CAM\citep{selvaraju2017gradcam} to visualize the critical image regions corresponding to the caption input.
The Grad-CAM is computed on the average self-attention maps in the vision encoder's last layer, where gradients are acquired by the contrastive loss $\mathcal{L}_{itc}$.
The results of the CLIP-ViT-L/14 (FT-L14), CLIP-ViT-B/32 (FT-B32), SE-CLIP$_V$, and MoPE-CLIP$_V$ are shown in Figure~\ref{fig:cam_diff_models}. 
We observe that the visualizations from FT-L14 are more precise than FT-B32. 
The FT-L14 model has a smaller patch size of 14 and thus locates more detailed regions, like the ``frisbee'' in the middle example.
Additionally, MoPE-CLIP$_V$ can effectively capture some important regions like FT-L14.
Both FT-B32 and SE-CLIP$_V$ miss the ``a white bus'' in the top example while losing ``one cat'' in the bottom example.
MoPE-CLIP$_V$ captures these important objects correctly.
This indicates that our proposed MoPE-CLIP$_V$ provides fruitful information for the retrieval task.

% \vspace{-0.5cm}
\section{Conclusion}
In this paper, we investigate diverse methods to downsize VLP models and focus on exploring better pruning solutions.
We propose the Module-wise pruning error (MoPE) metric, offering an accurate measure of the CLIP module's importance.
On top of the MoPE metric, we introduce a unified framework and an advanced distillation loss for structured pruning during the pre-training and fine-tuning stages.
Extensive experiments have demonstrated that our MoPE-CLIP achieves surprising success across various downstream tasks.

{
    \small
    \bibliographystyle{ieeenat_fullname}
    \bibliography{main}
}

\clearpage
\setcounter{page}{1}
\maketitlesupplementary

\appendix

\renewcommand\thefigure{\Alph{section}\arabic{figure}}
\renewcommand\thetable{\Alph{section}\arabic{table}}
\setcounter{figure}{0}
\setcounter{table}{0}

\section{Related work} 
\label{sec:appx_related}

\paragraph{Vision-Language Pre-trained Models.} 

Benefiting from the efficiency of contrastive learning, vision-language pre-trained models like \citep{radford2021clip,jia2021ALIGN,yuan2021florence,yao2021filip,zhai2022lit,li2021albef,zhili2022task,dai2023instructblip,gou2023mixture} have achieved advanced capability across downstream tasks.
Such dual-stream models have efficient inference speed on multi-modal tasks like retrieval, as the image/text features can be computed offline \citep{yao2021filip,dai2022enabling}.
However, these models are often pre-trained with millions or billions of image-text pairs from scratch, which is computationally expensive \citep{cherti2023openclip,shi2023upop,shi2023crossget}.
Later works \citep{mu2022slip, li2021declip, lee2022uniclip} propose to use more complex objectives to reduce the amount of pre-training data. Others \citep{yang2023alip, kim2023mcd} intend to reduce the influence of noisy and unmatched image-text pairs.
However, these methods lead to less competitive retrieval performance.
In this work, we 
show that we can prune the original pre-trained CLIP to a desired size and significantly lift up the performance of the pruned model in a data-efficient way, i.e., with several magnitudes fewer pertaining data than the original CLIP.

\paragraph{Pruning of Transformer-based Models.}
Various methods have been proposed to compress uni-modal vision and language transformer models \citep{wang2019dbp,michel2019sixteenheads,voita2019analyzing_attn,lagunas2021blockpruning,chen2021SviTE, zhang2023epitopological, liu2024task, zhang2023efficient}.
Among them, structured pruning methods remove unimportant structured components 
(e.g., attention heads, FFN neurons, and Transformer layers) in the network.
Depending on how the pruned components are determined, 
pruning methods could be divided into two categories: search-based and metric-based methods.
Search-based methods \citep{chavan2022vitslimming,tao2023structuredgpt} usually apply masks on the structured components 
and need a searching process to determine their importance.
On the other hand, metric-based methods apply various metrics to determine module importance and result in a single-shot pruning process.
Widely used metrics include the magnitude of weight \citep{han2015learning,he2018soft,zhu2017prune} and the variant in loss \citep{molchanov2016cnn_loss, michel2019sixteenheads, molchanov2019importance_estimation}.
Some researchers \citep{fan2019reducingbertdepth, sajjad2020poormanbert} explore different strategies for pruning BERT layers, such as  ``every other'', ``bottom or top dropping'' and ``search on valid'' like CNN Oracle Filter Pruning \citep{molchanov2016cnn_filter,abbasi2017cnn_Oracle}.
% Interestingly, ``every other'' is proven to work surprisingly well, and DynaBERT \citep{hou2020dynabert} directly adopts this to produce dynamic depth networks. 
Notably, the ``every other" strategy has been proven effective \citep{fan2019reducingbertdepth,sajjad2020poormanbert}, with DynaBERT \citep{hou2020dynabert} implementing it to create dynamic depth networks. 
Additionally, pruning is often used in combination with knowledge distillation, which 
transfers knowledge from the original unpruned teacher model to the smaller pruned model with different kinds of knowledge \citep{sun2020mobilebert,sanh2020movement_pruning,xia2022cofi}.

In contrast to the extensive research on compressing unimodal Transformer-based models, compression of multimodal models remains under-explored.
Our experiments show that directly using widely-used metrics \citep{han2015learning,michel2019sixteenheads} or ``every other'' strategy \citep{sajjad2020poormanbert,fan2019reducingbertdepth} for VLP pruning leads to unsatisfactory performance, indicating the demand for exploring more accurate metrics to measure module importance of VLP models across multi-modal tasks.
% cannot accurately reflect the module importance across multi-modal tasks.
Recently, EfficientVLM \citep{wang2022efficientvlm} proposes to distill the VLP model in the pre-training stage and then prune attention heads during the task-specific fine-tuning stage, but the distillation stage proved not optimal in our experiments.
Another work Upop \citep{shi2023upop} uses a unified and progressive search-based pruning method on vision-language models, but the search process is expensive and is hard to apply to the pre-training stage.
TinyCLIP \citep{wu2023tinyclip} proposes a multi-stage pruning and distillation method for pre-training small OpenCLIP models \citep{cherti2023openclip}. However, the design of the multi-stage is complex and the final performance relies on the huge pre-training dataset LAION400M \citep{schuhmann2021laion}.
In this work, we propose a simple but effective metric called MoPE, which serves as a general importance measure of various compressible components like attention heads, FFN neurons, and Transformer layers.
% which can be applied to both the pre-training and fine-tuning stages.
Based on MoPE metric, we design a unified pruning framework applied to both the pre-training and fine-tuning stages, resulting in state-of-the-art MoPE-CLIP models.

\section{Implementation Details} 
\label{sec:appx_implementation}

\subsection{Detailed Experimental Settings} \label{subsec:appx_train_detail}

Here we describe the detailed setups for our experiments of two compression stages. 
For all experiments, we use the same random seed (e.g., 42) to ensure reproduction.

\paragraph{Details for Evaluation Benchmarks.}
For retrieval tasks, we split the MSCOCO \citep{lin2014mscoco} and Flickr30K \citep{plummer2015Flickr30K} datasets following \citep{karpathy2015dataset}.
For classification tasks, we adopt 11 downstream datasets following \citep{zhang2021tip-adapter,zhang2023prompt}, 
including CIFAR10, CIFAR100 \citep{krizhevsky2009cifar10}, Caltech101 \citep{fei2004Caltech101}, Flowers102 \citep{nilsback2008flowers}, Oxford Pets \citep{parkhi2012cats}, DTD \citep{cimpoi2014dtd}, Stanford Cars \citep{krause2013cars},  
FGVC Aircraft \citep{maji2013aircraft}, SUN397 \citep{xiao2010sun397}, Food101 \citep{bossard2014food} and ImageNet \citep{deng2009imagenet}.

\paragraph{Details for Fine-tuning Stage Compression}
Table \ref{tab:hyper_ft-kd} summarizes the hyperparameters for fine-tuning CLIP-ViT-L/14 and distilling CLIP-VIT-B/32.
During the distilling process, we first fix the model and train the linear layer for 5 epochs with a learning rate of 1e-5 to learn a better mapping function.
Table \ref{tab:hyper_prune} lists the detailed retraining setups for MagnCLIP, DynaCLIP, MoPE-CLIP, and SE-CLIP in image-to-text retrieval (TR) and text-to-image retrieval (IR) tasks
The text encoders of these models are fixed for the TR task, while image encoders are frozen for the IR task.
For SE-CLIP, we add a linear layer to align feature space, and the hidden distillation loss is excluded due to the unmatched number of image patches.

% hyperparameters for fine-tuning
\begin{table}[!t]
\resizebox{\linewidth}{!}{
\begin{tabular}{l|cc}
\toprule
Config              & Fine-tuning    &   Distilling                         \\ 
\midrule
Optimizer          & \multicolumn{2}{c}{AdamW,  $\beta=(0.9,0.98)$} \\
% Optimizer          & \multicolumn{2}{c}{AdamW}                                     \\
% Optimizer momentum & \multicolumn{2}{c}{$\beta_1$=0.9, $\beta_2$=0.98}             \\
LR schedule        & \multicolumn{2}{c}{CosineLRScheduler}                         \\
Weight decay       & \multicolumn{2}{c}{3e-4}                                      \\
Warmup ratio       & \multicolumn{2}{c}{0.1}                                       \\
% Eps                & \multicolumn{2}{c}{1e-6}                                     \\
Init LR            & 3e-6            & 1e-6                                        \\
Batch size         & 256             & 1024                                        \\
Training epochs    & 12              & 15                                          \\ 
% Distillation       & NA              & \multicolumn{1}{l}{$\lambda_1=\lambda_2=1$} \\ \hline
Distillation       & N/A              & $\mathcal{L}_{sim} + \mathcal{L}_{feat}$  \\ 
\bottomrule
\end{tabular}
}
\caption{Experimental setup for fine-tuning CLIP-VIT-L/14 or distilling CLIP-ViT-B/32. }
\label{tab:hyper_ft-kd} 
% \vspace{-0.5em}
\end{table}

% hyperparameters for pruning
\begin{table}[t]
\resizebox{\linewidth}{!}{
\begin{tabular}{l|cc}
\toprule
Downstream Task     & Image-to-text          & Text-to-image            \\ 
\midrule
Optimizer          & \multicolumn{2}{c}{AdamW,  $\beta=(0.9,0.98)$} \\
% Optimizer          & \multicolumn{2}{c}{AdamW}                         \\
% Optimizer Momentum & \multicolumn{2}{c}{$\beta_1$=0.9, $\beta_2$=0.98} \\
LR schedule        & \multicolumn{2}{c}{CosineLRScheduler}             \\
Weight decay       & \multicolumn{2}{c}{3e-4}                          \\
Warmup ratio       & \multicolumn{2}{c}{0.1}                           \\
% Eps                & \multicolumn{2}{c}{1e-6}                          \\
Init LR            & 2e-5                   & 8e-5                     \\
Batch size         & 256                    & 1024                     \\
Training epochs    & 20                     & 10                       \\ 
\bottomrule
\end{tabular}
}
\caption{Experimental setup for retraining MagnCLIP, DynaCLIP, MoPE-CLIP and SE-CLIP across TR and IR tasks.}
\label{tab:hyper_prune} 
% \vspace{-0.5em}
\end{table}

% hyperparameters for pretraining
\begin{table}[t]
\resizebox{\linewidth}{!}{
\begin{tabular}{l|cc}
\toprule
Config            & Pre-training             & Further Fine-tuning            \\ 
\midrule
Optimizer          & \multicolumn{2}{c}{AdamW,  $\beta=(0.9,0.98)$} \\
LR schedule        & \multicolumn{2}{c}{CosineLRScheduler}             \\
Weight decay       & \multicolumn{2}{c}{3e-4}                          \\
Warmup ratio       & 0.02                     & 0.1                    \\
% Eps                & \multicolumn{2}{c}{1e-6}                          \\
Init LR            & 5e-5                     & 4e-5                   \\
Batch size         & 512                      & 512                    \\
Training epochs    & 20                       & 15                     \\ 
\bottomrule
\end{tabular}
}
\caption{Experimental setup for pre-training DynaCLIP and MoPE-CLIP and further fine-tuning on downstream tasks.}
\label{tab:hyper_pretrain} 
\vspace{-1.em}
\end{table}

\paragraph{Details for Pre-training Stage Compression}
We list the detailed setup for pretraining stage compression in Table~\ref{tab:hyper_pretrain}. 
% DynaCLIP and MoPE-CLIP are pre-trained on CC3M and finetuned on MSCOCO or Flickr30K using the same hyperparameters.
MoPE-CLIP adopts the Recall Mean on MSCOCO validation dataset as the specific MoPE metric.
DynaCLIP and MoPE-CLIP share the same hyperparameters.

\subsection{Main Algorithm}

We illustrate the computation process of the MoPE metric in Algorithm \ref{alg:mope}, and our unified pruning framework resulting in MoPE-CLIP in Algorithm \ref{alg:main}.

% We illustrate our unified pruning framework for both pre-training and fine-tuning stage compression in Algorithm \ref{alg:main}.

\begin{algorithm}[h]
\small
\caption{Module-wise Pruning Error Metric}
\label{alg:mope}
\vspace{0.3ex}
\hspace*{0.02in} {\bf Input:} CLIP model $\ms{f_{\varphi }}$, Module $\ms{\theta}$, Dataset $\ms{\mathcal D}$ \\
%算法的输入， \hspace*{0.02in}用来控制位置，同时利用 \\ 进行换行
\hspace*{0.02in} {\bf Output:} Importance of $\ms{\theta}$ %算法的结果输出
\begin{algorithmic}[1]
\Procedure{MoPE\ }{$\ms{{f_{\varphi }}, \theta , \mathcal D}$}:
\vspace{0.3ex}
    \State{Compute the full CLIP Performance on $\ms{\mathcal D}$: $\mathcal{Z}\left [ f_{\varphi}\right ]$}
    \State{Compute the CLIP$_{\ms{\theta=0}}$ Performance on $\ms{\mathcal D}$: $\mathcal{Z}\left [f_{\varphi-\theta} \right ]$}
    % \State{Compute the MoPE$_{\ms{\theta}}$ by Equation \ref{eq:MoPE}}
    \State{Compute the MoPE$_{\ms{\theta}} = \mathcal{Z}\left [ f_{\varphi}\right ] - \mathcal{Z}\left [f_{\varphi-\theta} \right ]$}
  \vspace{0.3ex}
	\State{\textbf{return} MPWE$_{\ms{\theta}}$}
\EndProcedure
\vspace{-0.2em}
\end{algorithmic}
\end{algorithm}

\begin{algorithm}[h]
\caption{MoPE-CLIP: Pruning with MoPE Metric}
\small
\label{alg:main}
\vspace{0.3ex}
\hspace*{0.02in} {\bf Input:} CLIP model ${f_{\varphi }}$, Validation Set $\mathcal D_{val}$, Training Set $\mathcal D_{train}$
\hspace*{0.02in} {\bf Output:} MoPE-CLIP model %算法的结果输出
\begin{algorithmic}[1]
\vspace{0.3ex}
    % \vspace{0.3ex}
    \State{Partition the Attention Heads in $N \times L$ modules}
    \For{$l$ in $1,\ ..., L$}
        \For{head $h$ in $1,\ ..., N$}
                \State{$\triangleright$\ \ \textit{run in parallel}}
                \vspace{0.15ex}
                \State{MoPE$\ms{_h}$ $\leftarrow$ MoPE($f_{\varphi }$, $h$,  $\mathcal D_{val}$)}
                \vspace{0.15ex}
                \State{Update $\mathcal{C}_{head}$}
        \EndFor
    \EndFor
    \vspace{0.5ex}
    \State{CLIP $f_{\varphi }'$ $\leftarrow$ Rewire Neurons in FFN by gradient}
    \vspace{0.3ex}
    \State{Partition the FFN Neurons in N groups}
    \vspace{0.1ex}
    \For{group $n$ in $1,\ ..., N$}
        \State{$\triangleright$\ \ \textit{run in parallel}}
        \vspace{0.15ex}
        \State{MoPE$\ms{_n}$ $\leftarrow$ MoPE($f_{\varphi }'$, $n$, $\mathcal D_{val}$)}
        \vspace{0.15ex}
        \State{Update $\mathcal{C}_{neuron}$}
    \EndFor
    \vspace{0.5ex}
    \If{Compression in fine-tuning stage}
        \vspace{0.35ex}
        \State{MoPE-CLIPw $f_{Cw}$ $\leftarrow$ Prune the CLIP in width} 
        \Statex{\qquad \qquad \qquad \qquad \qquad \quad and retrain on $\mathcal D_{train}$}
        \For{layer $l$ in $1,\ ..., L$}
            \State{$\triangleright$\ \ \textit{run in parallel}}
            \vspace{0.15ex}
            \State{MoPE$\ms{_l}$ $\leftarrow$ MoPE($f_{Cw}$, $l$, $\mathcal D_{val}$)}
            \vspace{0.15ex}
            \State{Update $\mathcal{C}_{layer}$}
        \EndFor
        \State{MoPE-CLIP $\leftarrow$ Prune the MPEE-CLIPw in depth}
        \Statex{\qquad \qquad \qquad \quad \quad and retrain on $D_{train}$}
    \vspace{0.3ex}
    \ElsIf{Compression in pretraining stage}
        \For{layer $l$ in $1,\ ..., L$}
            \State{$\triangleright$\ \ \textit{run in parallel}}
            \vspace{0.15ex}
            \State{MoPE$\ms{_l}$ $\leftarrow$ MoPE($f_{\varphi }$, $l$, $\mathcal D_{val}$)}
            \vspace{0.15ex}
            \State{Update $\mathcal{C}_{layer}$}  
        \EndFor
        \vspace{0.1ex}
        \State{MoPE-CLIP $\leftarrow$ Prune the CLIP in width and depth} \Statex{\qquad \qquad \qquad \quad \quad and retrain on $D_{train}$}
    \EndIf
    \State{\textbf{return} the MoPE-CLIP}		

\end{algorithmic}
\end{algorithm}

% ============= C More experiments

\section{More Experimental Results}
\label{sec:appx_results}

\subsection{Detailed Comparison with Baselines}
\label{subsec:appx_comparison}

We provide a more detailed comparison with DynaCLIP$_V$, MagnCLIP$_V$, and UPop in the following.

\vspace{-0.5em}

\begin{figure}[h]
    \centering
    \includegraphics[width=0.7\linewidth]{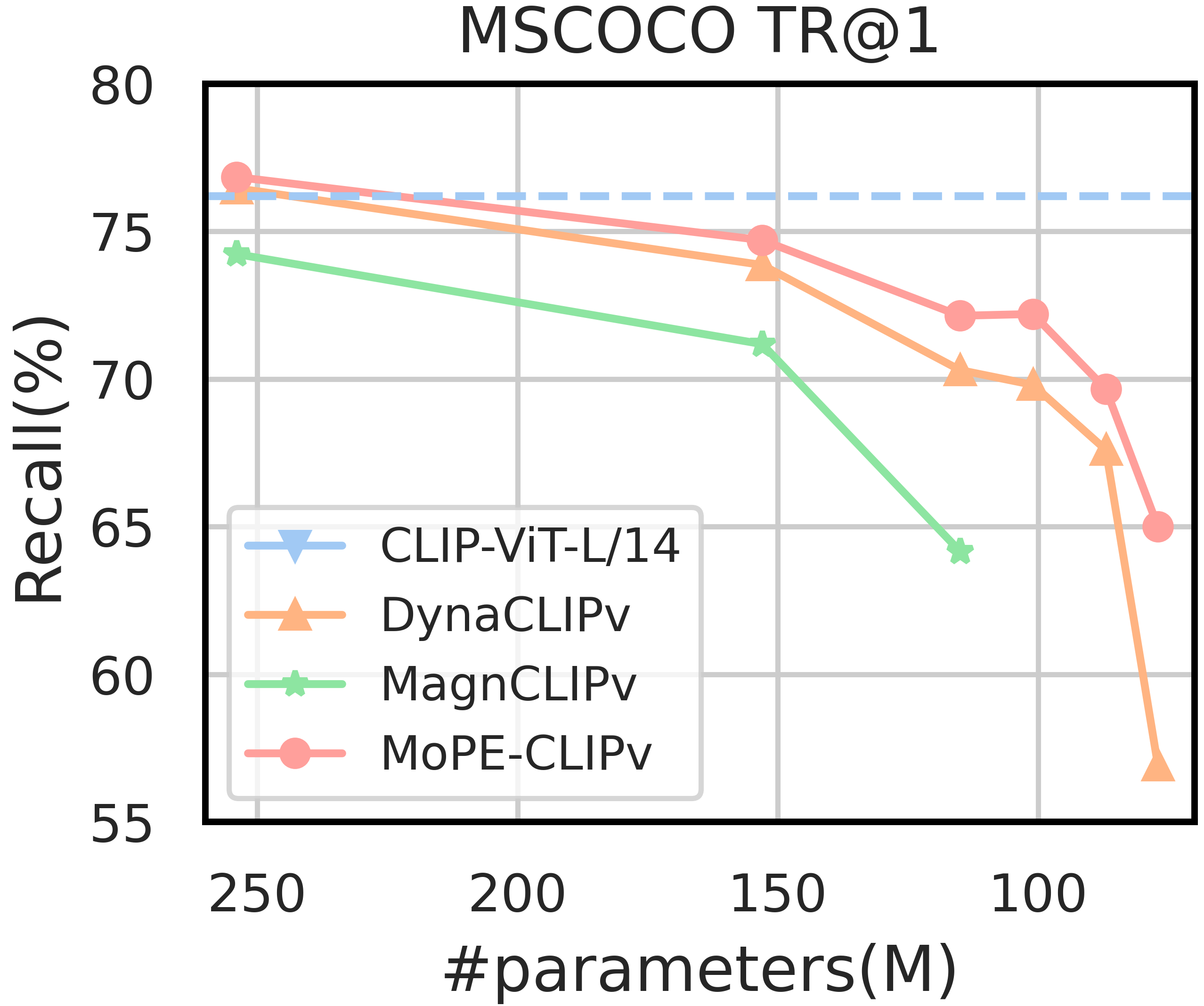}
    \vspace{-5pt}
    \caption{Comparsion of different pruning ratios.
    }
    \label{fig:pruning_impact}
    % \vspace{-0.5em}
\end{figure}

\paragraph{Pruning Ratios.}
To further evaluate our MoPE-CLIP performance under different model sizes.
We test six pruning ratios with the model performance plotted in Fig.~\ref{fig:pruning_impact}. 
MoPE-CLIP consistently stands above all other baselines (i.e., DynaCLIP and MagnCLIP) across different pruning ratios. The gap becomes even larger for higher sparsities.

\vspace{-5pt}
\begin{table}[h]
\centering
\resizebox{0.95\linewidth}{!}{
\begin{tabular}{l|l|cc}
\toprule
\multirow{2}{*}{Model} & \multirow{2}{*}{Params} & \multicolumn{2}{c}{MSCOCO (5K test set)}  \\
                       &                         & TR @1               & IR @1               \\ \midrule
UPop-Teacher           & 856M                    & 71.5                & 56.8                \\
UPop-CLIP              & 280M ${\color{red}\downarrow 67\%}$             & 56.1 ${\color{red}\downarrow 21\%}$         & 41.1 ${\color{red}\downarrow 27\%}$         \\
Upop-CLIP (+KD)        & 280M ${\color{red}\downarrow 67\%}$             & 58.6 ${\color{red}\downarrow 18\%}$         & 44.3 ${\color{red}\downarrow 22\%}$         \\ \midrule
MoPE-Teacher           & 390M                    & 76.2                & 58.8                \\
\rowcolor{orange!10}
MoPE-CLIP              & \textbf{122M ${\color{red}\downarrow 69\%}$}    & \textbf{70.7 ${\color{red}\downarrow 7\%}$} & \textbf{54.7 ${\color{red}\downarrow 7\%}$} \\ 
\bottomrule
\end{tabular}
}
% \vspace{-5pt}
\caption{UPop and MoPE-CLIP on MSCOCO.
}
\label{tab:upop} 
\vspace{-1.em}
\end{table}

\paragraph{Relative Comparison with UPop.}
We further compare Upop with Knowledge Distillation in Tab.~\ref{tab:upop}. 
MoPE-CLIP is superior to Upop (+KD) both on the relative performance drop and absolute task score, given a comparable relative decrease ($69\%$ vs $67\%$) in the number of parameters. 
Moreover, MoPE-CLIP's advantage is notable, as compressing smaller original model sizes is more challenging.

\subsection{Fine-tuning Stage Compression on Flickr30K} 
\label{subsec:appx_flickr}

To demonstrate the robustness of the MoPE metric across different data distributions, we further evaluate MoPE-CLIP on Flickr30K Dataset during fine-tuning stage compression.

\vspace{-0.2em}

\paragraph{Results for Image-to-text Retrieval.}
Following the setting in 
% Section \ref{subsec:exp_finetuning}, 
Section \rred{4.1},
we compress the vision encoder of fine-tuned CLIP-ViT-L14 (FT-L14) for image-to-text retrieval. 
% Since the pruned model performs best of the three architectures, 
We mainly compare the fine-tuned performance of our MoPE-CLIP$_V$ with fine-tuned CLIP-ViT-B/32 (FT-B32), DynaCLIP$_V$, and UPop-CLIP \cite{shi2023upop}
on the Flickr30K dataset.
In particular, we compute the loss gradient and MoPE metric (TR Mean) in Flickr30K  \citep{plummer2015Flickr30K} validation dataset for DynaCLIP$_V$ and MoPE-CLIP$_V$.
The results are presented in Table \ref{tab:vision_compress_flickr}.
We could observe that once depth pruning is added to DynaCLIP$_V$, the TR@1 drops from 89.6\% to 84.5\%, while the MoPE-CLIP$_V$ with 87M vision encoder maintains competitive retrieval and surpasses the FT-B32.
In addition, our MoPE-CLIP$_V$ with 115M vision encoder termed an entire model of 234M parameters outperforms the UPop-CLIP with 280M parameters by 8.2\% TR@1.
These results indicate the superiority of the MoPE metric across different downstream datasets.

\begin{table}[t]
\resizebox{\linewidth}{!}{
\begin{tabular}{c|ccc|ccc}
\toprule
\multirow{2}{*}{Approach} & \multicolumn{3}{c|}{Vision Encoder} & \multicolumn{3}{c}{Flickr30K (1K test set)}      \\ %\cline{2-6} 
                                   & Wdith      & Depth         & Parmas & TR@1           & TR@5           & TR@10          \\ 
\midrule
Teacher Model                     & 1024       & 24            & 304M   & 96.3          & 99.8          & 100.0         \\ 
\midrule
CLIP-ViT-B/32                         & 768        & 12          & 88M    & 87.7          & 97.7          & 99.3          \\ 
\midrule
\multirow{3}{*}{DynaCLIP$_V$ \citep{hou2020dynabert}}    & 512    & 24     & 153M   & \textbf{92.7}          & 99.4          & 99.8          \\
                                   & 384        & 24           & 115M   & 89.6          & 98.5          & 99.4          \\
                                   & 384        & 18           & 87M    & 84.5          & 97.3          & 98.5          \\ 
\midrule
\multirow{2}{*}{UPop-CLIP \citep{shi2023upop}}      
                            & N/A        & N/A       & 474M$^{\ddag}$       & \textbf{93.2}        & 99.4       & 99.8        \\                        
                            & N/A        & N/A        & 280M$^{\ddag}$      & 82.9        & 95.7       & 97.8        \\ 
\midrule
\multirow{3}{*}{MoPE-CLIP$_V$}     & 512        & 24            & 153M   & \textbf{92.7} & \textbf{99.5} & \textbf{99.9} \\
                                   & 384        & 24            & 115M   & \textbf{91.1} & \textbf{98.9} & \textbf{99.7} \\
                                   & 384        & 18            & 87M    & \textbf{88.5} & \textbf{98.5} & \textbf{99.6} \\ 
\bottomrule
\end{tabular}
}
\caption{Image-to-text retrieval results on the Flickr30K dataset.
The Params labeled as $^{\ddag}$ denote the parameters of the entire model.
}
 % ($m_w$,$m_d$)
\label{tab:vision_compress_flickr} 
% \vspace{-0.5em}
\end{table}

\begin{table}[t]
\resizebox{\linewidth}{!}{
\begin{tabular}{l|ccc|ccc}
\toprule
\multirow{2}{*}{Approach}       & \multicolumn{3}{c|}{Text Encoder} & \multicolumn{3}{c}{Flickr30K (1K test set)}   \\
                               & Width     & Depth     & Params    & IR @1         & IR @5         & IR @10        \\ 
\midrule
Teacher Model                  & 768       & 12        & 85M       & 84.7          & 97.4          & 99.0          \\ 
\midrule
CLIP-ViT-B/32                  & 512       & 12        & 38M       & 74.7          & 93.4          & 96.9          \\ 
\midrule
\multirow{2}{*}{DynaCLIP$_T$ \citep{hou2020dynabert}}  & 384       & 12        & 42M       & 84.1          & 97.1          & 98.7          \\
                               & 192       & 12        & 21M       & 80.3          & 95.7          & 98.0          \\ 
\midrule
\multirow{2}{*}{MoPE-CLIP$_T$} & 384       & 12        & 42M       & \textbf{85.1} & \textbf{97.4} & \textbf{99.1} \\
                               & 192       & 12        & 21M       & \textbf{83.5} & \textbf{97.2} & \textbf{98.8} \\ 
\bottomrule
\end{tabular}
}
\caption{Text-to-image retrieval results on the Flickr30K dataset. 
Pruning is applied in the width direction.
% The Params don't calculate Embedding layer parameters.
}
\label{tab:text_compress_flickr} 
\vspace{-0.5em}
\end{table}

\paragraph{Results for Text-to-image Retrieval.}
We compress the text encoder of fine-tuned CLIP-ViT-L/14 for text-to-image retrieval.
The pruning and retraining remain the same as the setting on the MSCOCO dataset and the results are illustrated in Table \ref{tab:text_compress_flickr}.
The MoPE-CLIP$_T$ exhibits significant performance on the Flickr30K dataset. 
Even at a 4x compression ratio, the MoPE-CLIP$_T$ surpasses the FT-B32 by 8.8\% IR@1 and DynaCLIP$_T$ by 3.2\% IR@1.
These superior results demonstrate that our MoPE-CLIP$_T$ provides a powerful text encoder for the text-to-image retrieval task.

\subsection{Further Discussion.} \label{subsec:appx_analysis_pruing}

\paragraph{Similarity matrix indicates pruning is the best architecture.}

\begin{figure}[t]
    \centering
    \includegraphics[width=1\linewidth]{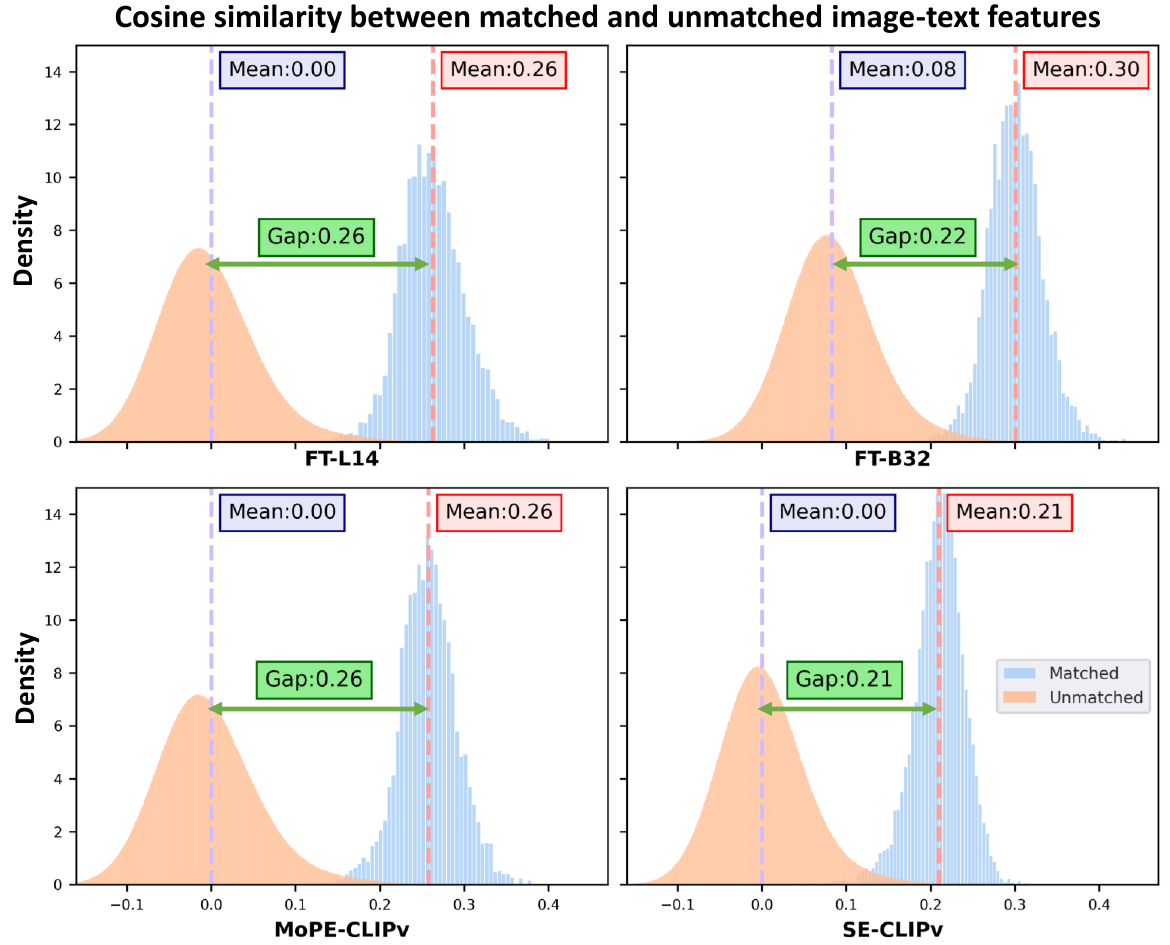}
    \caption{Histograms of cosine similarities between matched and unmatched image-text features. The green box represents the similarity gap.
    MoPE-CLIP$_V$ preserves a similar space to FT-L14.
    }
    \label{fig:similarity_matrix}
    \vspace{-1em}
\end{figure}

We compare and analyze the similarity matrix of three architectures discussed in 
% Section \ref{sec:preliminary} 
Section \rred{2}
since it directly influences retrieval performance.
In particular, we sample 5k image-text pairs from the MSCOCO \citep{lin2014mscoco} validation dataset and calculate the similarities between matched image-text features and unmatched pairs, as done in previous works \citep{udandarao2022sus, zhu2023APE}. 
Following \citep{baldrati2022conditioned_retrieval}, we suppose that the retrieval performance is more influenced by the similarity gap between matched and unmatched features. 
We compare the MoPE-CLIP$_V$ with fine-tuned CLIP-ViT-L/14 (FT-L14), fine-tuned CLIP-ViT-B/32 (FT-B32) and SE-CLIP$_V$.
From Figure \ref{fig:similarity_matrix}, we observe that FT-L14 has a larger gap between two similarities compared with FT-B32, reflecting its powerful performance.
The pruned MoPE-CLIP$_V$ shows a similar distribution and gap to FT-L14, while the SE-CLIP$_V$ even closes the gap, indicating the performance difference among these models.
Therefore, MoPE-CLIP$_V$, which preserves a similarity space like FT-L14, emerges as the best compact model architecture.

\begin{figure}[t]
    \centering
    \includegraphics[width=1\linewidth]{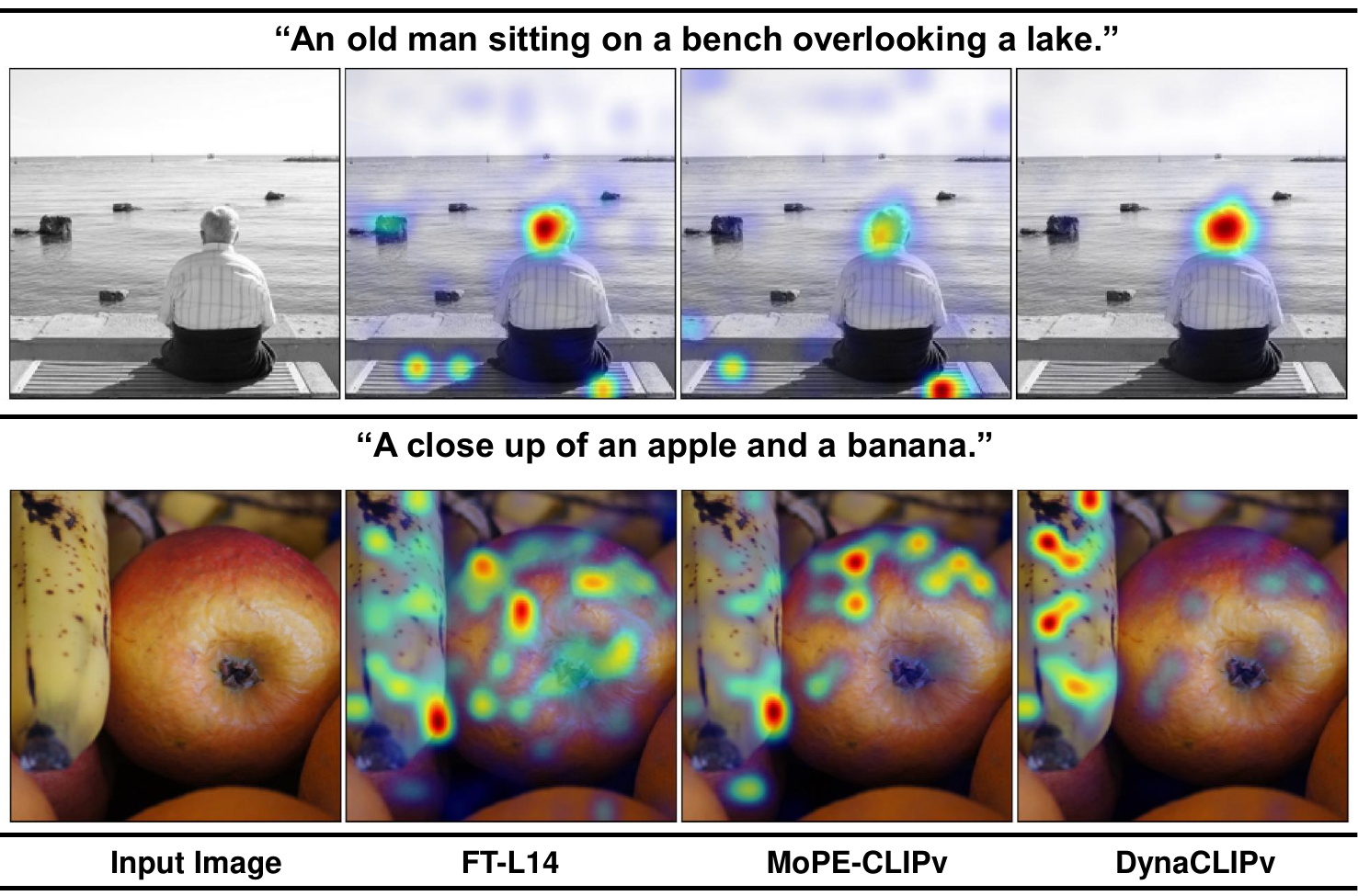}
    \caption{Grad-CAM visualization on the self-attention maps corresponding to the caption input.}
    \label{fig:cam_avg}
    \vspace{-0.5em}
\end{figure}

\begin{figure}[!t]
    \centering
    \includegraphics[width=1\linewidth]{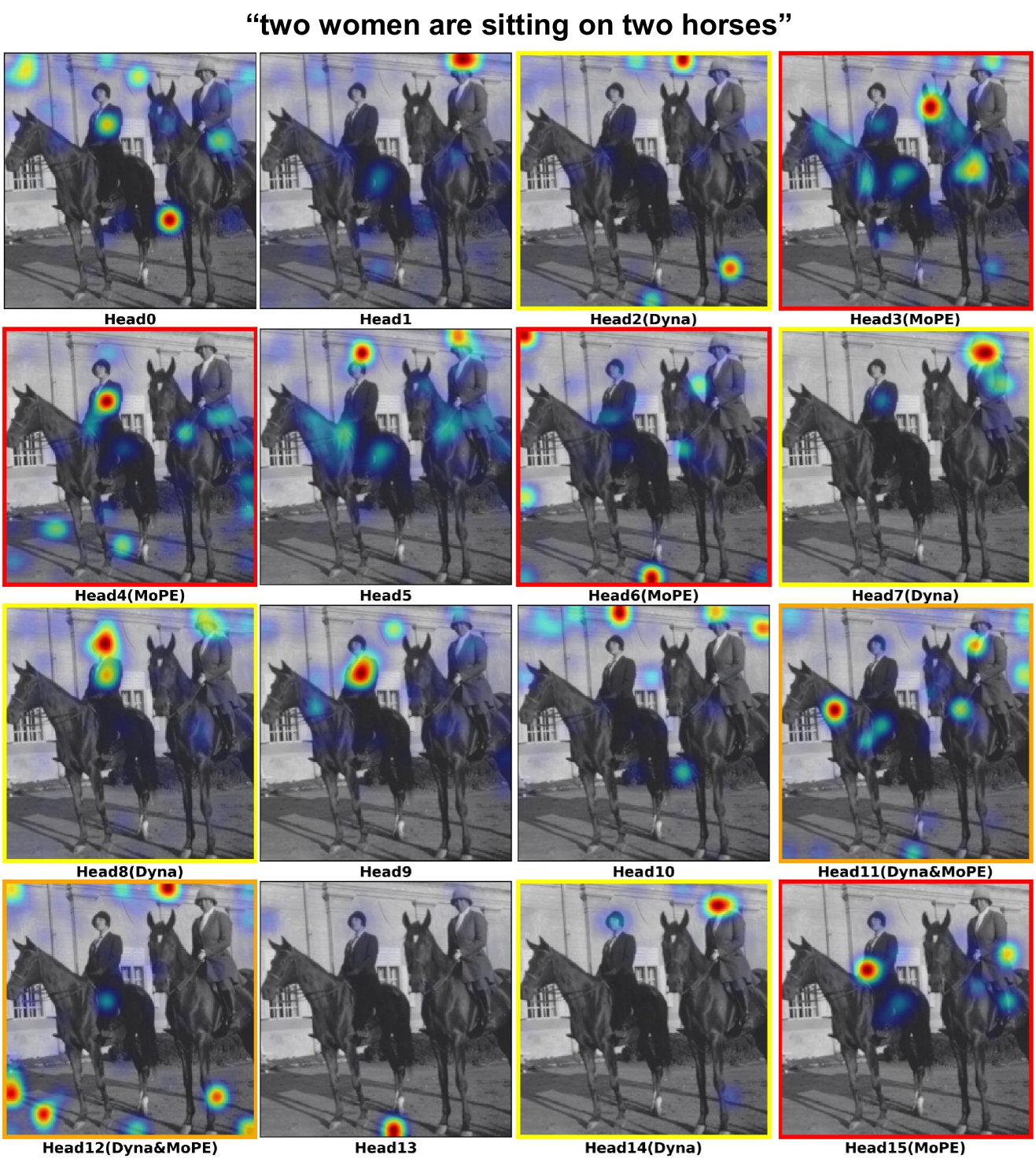}
    \caption{Grad-CAM visualization of the last layer self-attention maps for original FT-L14's vision encoder. Red box denotes preserved heads based on MoPE-CLIP$_V$. Yellow box denotes preserved heads based on DynaCLIP$_V$. Orange box denotes the head is preserved by two models simultaneously.}
    \label{fig:cam_each_head}
    \vspace{-1.2em}
\end{figure}

% fine-tuned retrieval evaluation
\begin{table*}[!t]
\begin{center}
\resizebox{\linewidth}{!}{
\begin{tabular}{l|cc|cc|c|cccccc|cccccc}
\toprule
\multirow{2}{*}{Method} & \multicolumn{2}{c|}{Vision Enocder} & \multicolumn{2}{c|}{Text Encoder} & Params(M) & \multicolumn{6}{c|}{MSCOCO (5K test set)}                                                     & \multicolumn{6}{c}{Flickr30K (1K   test set)}                                        \\ % \cmidrule{2-4}  \cmidrule{5-7}
                          & Width            & Depth            & Width           & Depth           & Vision + Text                           & TR @1         & TR @5         & TR @10        & IR @1         & IR @5         & IR @10        & TR @1         & TR @5         & TR @10        & IR @1         & IR @5         & IR @10        \\ 
\midrule
\color{gray}{\textit{Pre-trained on WIT-400M}} \\
CLIP-ViT-L/14 \citep{radford2021clip}            & 1024             & 24               & 768             & 12              & 304 + 85                   & 76.2          & 92.9          & 96.4          & 58.8          & 82.8          & 89.5          & 96.3          & 99.8          & 100.0         & 84.7          & 97.4          & 99.0          \\
CLIP-ViT-B/32 \citep{radford2021clip}            & 768              & 12               & 512             & 12              & 88 + 38                    & 66.2          & 87.7          & 92.8          & 49.4          & 75.8          & 84.7          & 87.7          & 97.7          & 99.3          & 74.7          & 93.4          & 96.9          \\
\midrule
\color{gray}{\textit{Pre-trained on CC3M}} \\

DynaCLIP$_{base}$ \citep{hou2020dynabert}           & 384              & 18               & 384             & 12              & 86 + 42                    & 70.7          & 90.0          & 94.6          & 53.8          & 80.5          & 87.9          & 90.0          & 98.8          & 99.7          & 79.0          & 95.5          & 97.9          \\
DynaCLIP$_{small}$ \citep{hou2020dynabert}          & 384              & 18               & 192             & 12              & 86 + 21                    & 69.3          & 89.5          & 94.5          & 52.3          & 79.1          & 87.1          & 89.4          & 98.1          & 99.7          & 77.3          & 95.0          & 97.4          \\
\rowcolor{orange!10} 
MoPE-CLIP$_{base}$        & 384              & 18               & 384             & 12              & 86 + 42                    & \textbf{71.9} & \textbf{91.4} & \textbf{95.7} & \textbf{54.9} & \textbf{81.1} & \textbf{88.6} & \textbf{92.1} & \textbf{98.8} & \textbf{99.0} & \textbf{80.6} & \textbf{95.6} & \textbf{98.1} \\
\rowcolor{orange!10} 
MoPE-CLIP$_{small}$       & 384              & 18               & 192             & 12              & 86 + 21                    & \textbf{71.2} & \textbf{90.9} & \textbf{95.0} & \textbf{53.7} & \textbf{80.5} & \textbf{87.9} & \textbf{90.8} & \textbf{98.6} & \textbf{99.6} & \textbf{79.3} & \textbf{95.5} & \textbf{97.9} \\

\midrule
\color{gray}{\textit{Pre-trained on YFCC15M}} \\
CLIP-ViT-B/32$^{\dag}$ \citep{radford2021clip}      & 768              & 12               & 512             & 12              & 88 + 38                    & 34.5          & 63.5          & 75.2          & 24.0          & 50.8          & 63.5          & 57.4          & 84.7          & 90.2          & 40.4          & 69.5          & 79.6          \\
SLIP-ViT-B/32$^{\dag}$ \citep{mu2022slip}           & 768              & 12               & 512             & 12              & 88 + 38                    & 43.7          & 71.8          & 82.4          & 31.0          & 58.8          & 70.3          & 68.9          & 91.9          & 95.1          & 51.0          & 79.5          & 86.8          \\
DeCLIP-ViT-B/32$^{\dag}$ \citep{li2021declip}       & 768              & 12               & 512             & 12              & 88 + 38                    & 47.9          & 75.5          & 84.6          & 33.8          & 62.7          & 71.4          & 73.6          & 93.9          & 97.2          & 55.9          & 83.4          & 90.2          \\
UniCLIP-ViT-B/32$^{\dag}$ \citep{lee2022uniclip}    & 768              & 12               & 512             & 12              & 88 + 38                    & 52.7          & 78.6          & 87.4          & 37.6          & 66.3          & 77.0          & 77.9          & 95.1          & 98.0          & 61.0          & 85.9          & 92.2          \\
MCD-ViT-B/32$^{\dag}$ \cite{kim2023mcd}             & 768              & 12               & 512             & 12              & 88 + 38                    & 55.6          & 81.2          & 89.5          & 38.2          & 67.4          & 78.5          & 79.3          & 95.2          & 98.0          & 63.1          & 87.2          & 92.3          \\
\rowcolor{orange!10} MoPE-CLIP$_{base}$        & 384                & 18               & 384             & 12              & 86 + 42     & \textbf{74.3} & \textbf{92.3} & \textbf{95.9} & \textbf{56.7} & \textbf{82.0} & \textbf{89.4} & \textbf{93.3} & \textbf{99.4} & \textbf{99.9} & \textbf{82.0} & \textbf{96.4} & \textbf{98.7}                 \\ 
\bottomrule
\end{tabular}
}
\end{center}
\vspace{-10pt}
\caption{
% Fine-tuned image-text retrieval results for pre-training stage compression.
Fine-tuned image-text retrieval results on MSCOCO and Flickr30K datasets.
DynaCLIP and MoPE-CLIP are pruned during the pre-training stage and further fine-tuned on downstream datasets.
% Our MoPE-CLIP$_{base}$ pre-trained on CC3M datasets outperforms the CLIP-ViT-B/32 pre-trained on WIT-400M on all the metrics.
$^{\dag}$ denotes the results are reported from \citep{lee2022uniclip,yang2023alip}.
}
% \vspace{-1em}
\label{tab:pretrain-ft}
\end{table*}

\begin{table*}[h]
\centering
\resizebox{\linewidth}{!}{
\begin{tabular}{l|cc|cc|c|ccc|cc|cc}
\toprule
\multicolumn{1}{l|}{\multirow{2}{*}{Method}}      & \multicolumn{2}{c|}{Vision Enocder} & \multicolumn{2}{c|}{Text Encoder} & Params (M)      & \multicolumn{3}{c|}{Training Details}     & \multicolumn{2}{c|}{MSCOCO} & \multicolumn{2}{c}{Flickr30K} \\
\multicolumn{1}{c|}{}                               & Width            & Depth            & Width           & Depth           & Vision + Text & Dataset  & GPU       & Batch size & TR @1        & IR @1        & TR @1         & IR @1         \\
\midrule
OpenCLIP \citep{cherti2023openclip} 
                & 12               & 12               & 8               & 12              & 88 + 39       & LAION-2B & 176x A100 & 33792      & 59.4         & 42.4         & 86.2          & 69.8          \\
TinyCLIP \citep{wu2023tinyclip}    
                & N/A              & N/A              & 8               & 6              & 39 + 19       & YFCC15M  & 32x A100  & 4096       & 54.9         & 38.9         & 84.4          & 66.7          \\
\rowcolor{orange!10}
MoPE-CLIP                       
                & 6                & 12               & 4               & 12             & 43 + 19        & YFCC15M   & 8x V100  & 1024           
                & \textbf{56.2} & \textbf{39.4} & \textbf{84.5} & \textbf{67.4} \\
\bottomrule
\end{tabular}
`}
\caption{
    Zero-shot image-text retrieval results of TinyCLIP and MoPE-CLIP. 
    The original model is OpenCLIP-ViT-B/16 pre-trained on the LAION-2B dataset.
}
\label{tab:openclip} 
\vspace{-1.em}
\end{table*}

\vspace{-1em}
% Visualization demonstrates MoPE-CLIP preserves more important heads.
\paragraph{Grad-CAM demonstrates MoPE-CLIP preserves more important heads.}

To better understand the effect of our MoPE metric, we use Grad-CAM \citep{selvaraju2017gradcam} to visualize the regions focused by DynaCLIP$_V$ and MoPE-CLIP$_V$. 
In detail, we select the model with a 115M vision encoder
and compute the Grad-CAM using self-attention maps averaged over all attention heads in the last layer of the vision encoder.
The gradients are acquired by contrastive loss $\mathcal{L}_{cont}$.
From Figure \ref{fig:cam_avg}, we could observe that the average attention map of MoPE-CLIP$_V$ is similar to original model (FT-L14), but the DynaCLIP$_V$ misses some important regions, like the "bench" in the top line and the "apple" in the bottom line.
Furthermore, We visualize the Gram-CAM of each head of the FT-L14 model and identify the preserved heads by DynaCLIP$_V$ or MoPE-CLIP$_V$. 
As shown in Figure \ref{fig:cam_each_head}, MoPE-CLIP$_V$ preserves heads 3, 4, and 15, which correspond to the crucial region of "sitting on the horse." Conversely, DynaCLIP$_V$ prunes these heads, leading to their exclusion.
This observation proves the precision of the MoPE metric in identifying and preserving vital information.

\paragraph{Width-and-depth pruning is preferred for pre-training compression.
}

Following 
% Section \ref{subsec:ablation}, 
Section \rred{4.3},
we extend our investigation to include both ``width-and-depth pruning" and ``width-first-then-depth pruning" strategies during the pre-training stage compression. 
We exclude the ``depth-first-then-width" strategy since it falls behind the "width-first-then-depth pruning" during the fine-tuning stage.
As indicated in Table \ref{tab:ablation_framework_pretrain},  ``width-first-then-depth pruning" shows superior performance.
However, the performance gap with ``width-and-depth pruning" narrows significantly compared to the fine-tuning stage. 
Notably, ``width-first-then-depth pruning" requires an additional 20 epochs in pre-training, which can be resource-intensive for many researchers. 
On the other hand, ``width-and-depth pruning" offers the dual benefits of one-stage pruning for faster training and the utilization of a larger set of image-text pairs, thereby yielding competitive performance. 
Consequently, we advocate for ``width-and-depth pruning" during the pre-training stage compression, as it strikes an optimal balance between training efficiency and model capability.

\begin{table}[t]
\centering
\resizebox{\linewidth}{!}{
\begin{tabular}{ll|cccc|cc}
\toprule
\multicolumn{2}{l|}{\multirow{2}{*}{Pruning Strategy}} & \multicolumn{2}{c}{MSCOCO} & \multicolumn{2}{c|}{Flickr30K} & \multicolumn{2}{c}{Training cost} \\
\multicolumn{2}{l|}{}                                  & TR @1        & IR @1       & TR @1          & IR @1         & Epochs         & GPU Hours        \\ 
\midrule
\multicolumn{2}{l|}{Width-and-depth}                   & 52.8         & 37.3        & 82.8           & 66.7          & 20             & 320              \\
\multicolumn{2}{l|}{Width-first-then-depth}            & 54.3         & 38.1        & 84.1           & 67.9          & 40             & 640              \\ 
\bottomrule
\end{tabular}
}
\caption{ Comprasion of retrieval performance and training cost in pruning 86M+42M MoPE-CLIP$_{base}$.}
\label{tab:ablation_framework_pretrain} 
\vspace{-1.5em}
\end{table}

\subsection{Fine-tuned Evaluation for Pre-training Stage} \label{subsec:appx_pretrain-ft}

As we discussed in 
% Section \ref{sec:preliminary}, 
Section \rred{2},
whether pruning during the pre-training stage and then fine-tuning outperforms pruning during the fine-tuning stage is an interesting question.
Therefore, we further fine-tune the DynaCLIP and MoPE-CLIP on downstream datasets and compare them with other baselines.
From Table \ref{tab:pretrain-ft}, we observe that the finetuned MoPE-CLIP and DynaCLIP exhibit significant performance on two datasets and enlarge the gap compared to fine-tuned CLIP-ViT-B/32.
This indicates that pruned models continually inherit the knowledge from the fine-tuned CLIP-ViT-L/14 during full fine-tuning.
Consequently, we compare the fine-tuned MoPE-CLIP$_{base}$ with MoPE-CLIP$_{V}$ in Table 
% \ref{tab:vision_small_models} 
\rred{1}
and find that 
% MoPE-CLIP$_{base}$ 
the former
showcases better TR@1.
% To answer the question, we find that fine-tuned MoPE-CLIP$_{base}$ outperforms the MoPE-CLIP$_{V}$ in Table by 2.2\% TR@1.
This indicates that pruning during the pre-training stage is more effective because more image-text pairs are included for learning, while the pruning during fine-tuning stage exhibits competitive results with much less training time.
In addition, if we enlarge the pre-training dataset to YFCC15M, fine-tuned UniCLIP \citep{lee2022uniclip} and MCD \citep{kim2023mcd} still fall short 
in comparison to MoPE-CLIP$_{base}$.
This aligns with the conclusion in
% stated in 
% Section \ref{subsec:exp_pretraining} 
Section \rred{4.2} 
% \textcolor[RGB]{197,74,61}{4.2}
that pruning offers a superior solution for obtaining compact VLP models.

\subsection{MoPE on OpenCLIP} \label{subsec:appx_openclip}

To assess our MoPE metric across various vision-language models, we adopted the setting used in TinyCLIP \citep{wu2023tinyclip} and further compressed the OpenCLIP-ViT-B/16 \citep{cherti2023openclip}, which is pre-trained on the LAION-2B dataset \citep{schuhmann2022laion5b}. 
Specifically, we prune both the vision and language encoders to half their original widths.
The MoPE metric is computed by Recall Mean on the MSCOCO validation dataset, following 
% Section \ref{subsec:exp_pretraining}
Section \rred{4.2}.
We then pre-train the reduced model on the YFCC15M dataset for 25 epochs, employing 16x NVIDIA V100 GPUs, and the results are presented in Table \ref{tab:openclip}.
We observe that our MoPE-CLIP, utilizing significantly fewer GPU resources, surpasses TinyCLIP in retrieval tasks on both MSCOCO and Flickr30K benchmarks, and narrows the performance gap with OpenCLIP.
However, due to limited computational resources, we were unable to increase the batch size to 4096 as done in TinyCLIP. Therefore, we anticipate further enhancements with the availability of more GPUs. 
These experiments validate the effectiveness of the MoPE metric across different VLP models and also demonstrate that our MoPE-CLIP offers a straightforward yet efficient approach for pre-training stage compression.

% {
%     \small
%     \bibliographystyle{ieeenat_fullname}
%     \bibliography{main}
% }

% \maketitle
% \input{sec/0_abstract}    
% \input{sec/1_intro}
% \input{sec/2_preliminary}
% \input{sec/3_method}
% \input{sec/4_experiment}
% \input{sec/5_conclusion}

\end{document}